\documentclass[sn-standardnature,iicol]{sn-jnl}% Standard Nature Portfolio Reference Style
\usepackage{rotating}

\usepackage{arydshln}
\jyear{2022}%

\theoremstyle{thmstyleone}%
\theoremstyle{thmstyletwo}%

\theoremstyle{thmstylethree}%

\begin{document}

\title[]{Root-aligned SMILES: A Tight Representation for Chemical Reaction Prediction}

\author[1]{\fnm{Zipeng} \sur{Zhong}}\email{zipengzhong@zju.edu,cn}
\author[2]{\fnm{Jie} \sur{Song}}\email{sjie@zju.edu,cn}
\author[2]{\fnm{Zunlei} \sur{Feng}}\email{zunleifeng@zju.edu,cn}
\author[3]{\fnm{Tiantao} \sur{Liu}}\email{liutiant@zju.edu.cn}
\author[1]{\fnm{Lingxiang} \sur{Jia}}\email{lingxiangjia@zju.edu,cn}
\author[1]{\fnm{Shaolun} \sur{Yao}}\email{yaoshaolun@zju.edu,cn}
\author[4]{\fnm{Min} \sur{Wu}}\email{swgcwumin@hdnewdrug.com}
\author*[3]{\fnm{Tingjun} \sur{Hou}}\email{tingjunhou@zju.edu.cn}
\author*[1]{\fnm{Mingli} \sur{Song}}\email{brooksong@zju.edu.cn}

\affil[1]{\orgdiv{College of Computer Science and Technology}, \orgname{Zhejiang University}, \orgaddress{\postcode{310027}, \state{Zhejiang}, \country{P.R. China}}}

\affil[2]{\orgdiv{School of Software Technology}, \orgname{Zhejiang University}, \orgaddress{\postcode{315048}, \state{Zhejiang}, \country{P.R. China}}}

\affil[3]{\orgdiv{Innovation Institute for Artificial Intelligence in Medicine of Zhejiang University, College of Pharmaceutical Sciences}, \orgname{Zhejiang University}, \orgaddress{\postcode{310058}, \state{Zhejiang}, \country{P.R. China}}}

\affil[4]{ \orgname{Hangzhou Huadong Medicine Group Pharmaceutical Research Institute}, \orgaddress{\postcode{310011}, \state{Zhejiang}, \country{P.R. China}}}

\abstract{
Chemical reaction prediction, involving forward synthesis and retrosynthesis prediction, is a fundamental problem in organic synthesis. A popular computational paradigm formulates synthesis prediction as a sequence-to-sequence translation problem, where the typical SMILES is adopted for molecule representations. However, the general-purpose SMILES neglects the characteristics of chemical reactions, where the molecular graph topology is largely unaltered from reactants to products, resulting in the suboptimal performance of SMILES if straightforwardly applied. In this article, we propose the root-aligned SMILES (R-SMILES), which specifies a tightly aligned one-to-one mapping between the product and the reactant SMILES for more efficient synthesis prediction. Due to the strict one-to-one mapping and reduced edit distance, the computational model is largely relieved from learning the complex syntax and dedicated to learning the chemical knowledge for reactions. We compare the proposed R-SMILES with various state-of-the-art baselines and show that it significantly outperforms them all, demonstrating the superiority of the proposed method.
}

\maketitle

Efficiently designing valid synthetic routes for valuable molecules plays a vital role in drug discovery and material design, which mainly involves forward synthesis prediction and retrosynthesis prediction.
The former predicts reaction outcomes~(product) with a given set of substrates~(reactants and reagents), and the latter predicts reactants for a target compound.
They are both challenging as the search space of all possible transformations is huge by nature. In the early days, expert synthetic chemists could design synthesis routes with their familiar reactions. To integrate more chemical knowledge and be more efficient, the first computer-aided synthesis planning program LHASA~\cite{pensak1977lhasa} was formally proposed by Corey \textit{et al.} and showed great potential. Since then, many rule-based organic synthesis systems have come out, such as SYNLMA~\cite{johnson1989designing}, WODCA~\cite{gasteiger2000computer}, and Synthia~\cite{szymkuc2016computer}. However, with the increase in chemical reaction rules, the cost of manually hard-coding chemical rules into computer systems is getting higher. 
Alternatively, people have begun to explore fully data-driven approaches, where the current literature can be roughly categorized into two schools: selection-based methods~ \cite{coley2017computer,segler2017neural,dai2019retrosynthesis,chen2021deep,guo2020bayesian,lee2021retcl} and generation-based methods~\cite{liu2017retrosynthetic,karpov2019transformer,zheng2019predicting,lin2020automatic,yan2020retroxpert,wang2021retroprime,tetko2020state,Seo_Song_Yang_Bae_Lee_Shin_Hwang_Yang_2021,kim2021valid,shi2020graph,somnath2021learning,sacha2021molecule}.
Selection-based methods turn synthesis prediction into a ranking or classification problem, where the goal is to rank the matched reaction templates~ \cite{coley2017computer,segler2017neural,dai2019retrosynthesis,chen2021deep} or target molecules~ \cite{guo2020bayesian,lee2021retcl} higher than those unmatched for the input molecule. 
Despite encouraging results achieved, selection-based methods are unable to predict templates that are not in the training set, which makes it suffer from poor generalization on new target structures and reaction types.
Generation-based methods, however, address the synthesis prediction with a generative model (\textit{e.g.}, transformers~\cite{liu2017retrosynthetic,karpov2019transformer,zheng2019predicting,lin2020automatic,yan2020retroxpert,wang2021retroprime,tetko2020state,Seo_Song_Yang_Bae_Lee_Shin_Hwang_Yang_2021,kim2021valid} or GNNs~\cite{shi2020graph,somnath2021learning,sacha2021molecule}) where target compounds are generated, which significantly alleviates the poor generalization issue of selection-based methods.

\begin{figure*}
    \centering
    \includegraphics[width=1.00\textwidth]{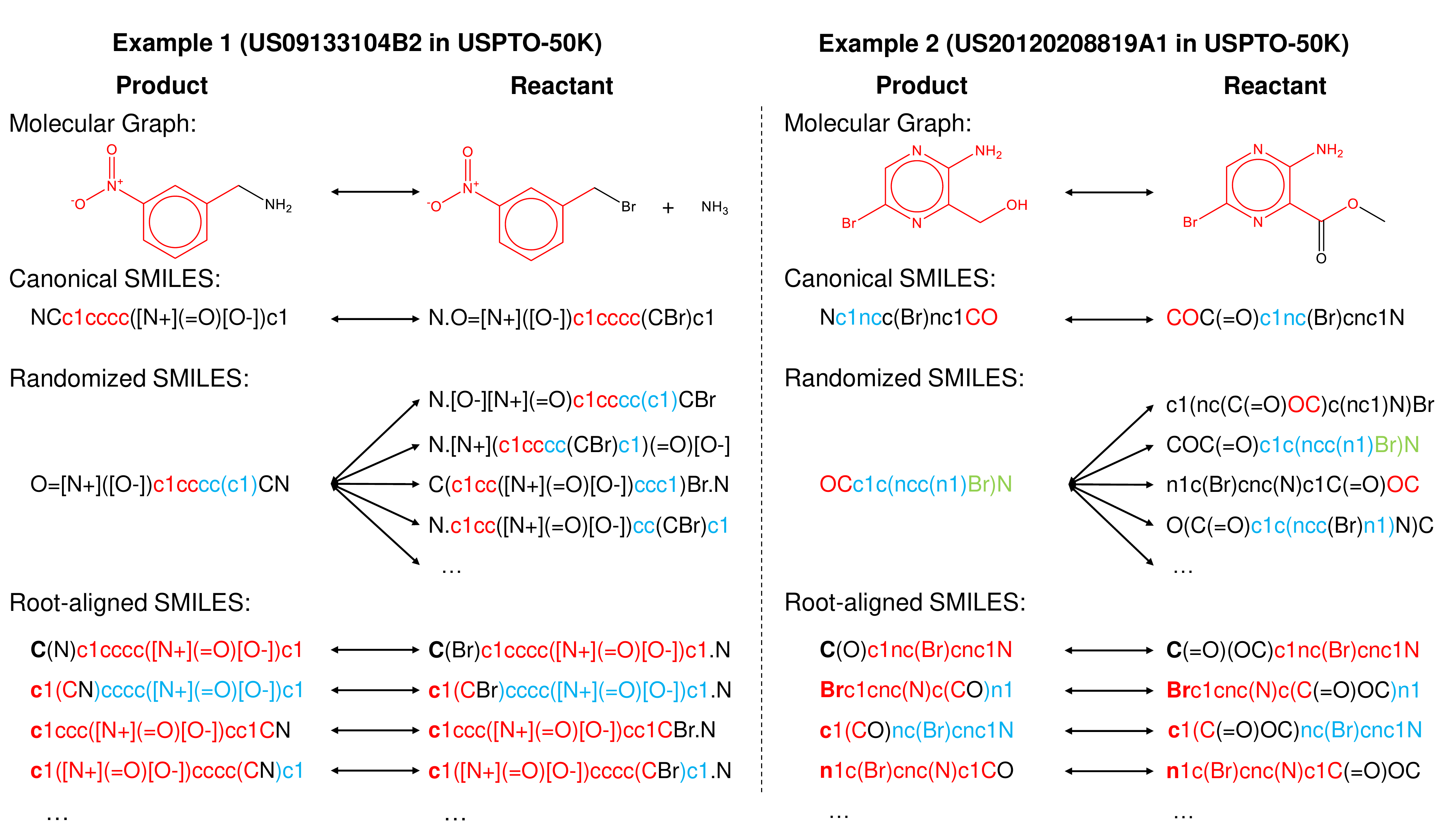}
    \caption{Comparison of differences between input and output with different molecular representations for retrosynthesis prediction.
    The root atom of root-aligned SMILES is bold.
    The common structures are represented with the same color.
    The more colored fragments in the output,
    the more similar they are.
    }
    \label{fig:R-SMILES}
\end{figure*}

Before applying generation-based methods for synthesis prediction, the first and critical step is to select the appropriate representation forms of both the product and the reactants. Two types of molecular representations are most widely used currently, including molecular graphs and string sequences. A molecular graph explicitly describes the topological structure of the molecule, upon which the recently well-developed GNNs~\cite{schlichtkrull2018modeling,velickovic2017graph} can be directly leveraged. However, graph-based representations involve a graph generation problem, which is challenging and usually solved by sequential graph edit operation predictions~\cite{shi2020graph,somnath2021learning,sacha2021molecule}. In contrast, another popular paradigm to represent molecules is using strings that are generated following some predefined chemical notation systems, of which the simplified molecular-input line-entry system (SMILES)~\cite{weininger1988smiles} is most widely used currently. With strings as the representations of molecules, synthesis prediction can be formulated as the typical seq2seq translation problem in natural language processing, where plenty of methods or models can be borrowed. 

SMILES has been widely used for both forward synthesis prediction~\cite{nam2016linking,schwaller2018found,schwaller2019molecular,tetko2020state} and retrosynthesis prediction~\cite{liu2017retrosynthetic,karpov2019transformer,zheng2019predicting,lin2020automatic,yan2020retroxpert,wang2021retroprime,tetko2020state,Seo_Song_Yang_Bae_Lee_Shin_Hwang_Yang_2021,kim2021valid} in the current literature. However, in this work, we argue that the general-purpose SMILES is deficient for the synthesis prediction problem. Since SMILES is generated by a depth-first traversal of the molecular graph, a molecule can have multiple valid SMILES representations, which leads to the existence of multiple correct output SMILES for a given input SMILES. The one-to-many mapping between input SMILES and output SMILES renders synthesis prediction extremely challenging as the computational model should learn not only the chemical rules for chemical reactions but also the SMILES syntax for SMILES string validity. Several canonicalization methods~\cite{o2012towards,schneider2015get} can be adopted to generate canonical SMILES that ensures a one-to-one mapping between molecules and SMILES. However, these methods are designed for each individual molecule without considering the relationship between product and reactant molecules, resulting in the large input-output SMILES discrepancy, as shown by the two examples (3-nitrobenzylamine and (3-Amino-6-bromo-2-pyrazinyl)methanol) in Fig.~\ref{fig:R-SMILES}. The large input-output SMILES discrepancy leaves the search space of reactants huge, degrading the performance of synthesis prediction models. Moreover, the canonical SMILES is incompatible with some data augmentation techniques where multiple SMILES are needed for one molecule to bypass the data scarcity issue, as the concept of ``canonical SMILES'' is violated by multiple SMILES for one molecule. 

In contrast to the large edit distance between the input and the output SMILES adopted in existing models, the molecular graph topology is in fact largely unaltered from reactants to products as the molecular changes usually occur locally during the chemical reactions~\cite{chen2021deep}. Therefore, in this article, we propose the root-aligned SMILES (R-SMILES) for more efficient synthesis prediction. As shown in Fig.~\ref{fig:R-SMILES}, for each chemical reaction, R-SMILES adopts the same atom as the root (\textit{i.e.}, the starting atom) of the SMILES strings for both the products and the reactants, which makes the input and the output SMILES maintain a one-to-one mapping and highly similar to each other.
The high similarity between the input and output makes synthesis prediction with R-SMILES very close to the typical autoencoding problem~\cite{pu2016variational,he2021masked} where the goal is to learn an identity mapping between the input and the output, with some bottleneck features summarizing the most important aspects in the data. Motivated by this, we propose a transformer-based autoencoder for synthesis prediction. With the proposed R-SMILES, we first pretrain the proposed autoencoder with the cheaply available unlabeled molecular data for extracting the compact molecular representations and mastering essential SMILES syntax in the decoder. 
Then the model is finetuned with the reaction data, where the model is largely relieved from learning the complex syntax and can be dedicated to learning the chemical knowledge for reactions. We conducted extensive experiments to validate the proposed method on various synthesis tasks, including product to reactant, product to synthon, synthon to reactant, and reactant to product which all demonstrates the efficiency of the proposed R-SMILES.
Compared with other baselines, our product-to-reactant and reactant-to-product variants both yield significantly superior performance on the public benchmark datasets.
For a better understanding of the proposed method, we visualize the cross-attention mechanism in transformer with R-SMILES. 
Furthermore, we provide several multistep retrosynthesis examples successfully predicted by our method, which illustrates its great potential in complicated synthesis planning tasks.

\begin{table*}
\small
\caption{\ An example (US20020192594A1 in USPTO-50K) of performing root alignment in the P2R stage. The root atoms are bold. (1) Select a reaction from the dataset. (2) Randomly select an atom as the root atom. [Cl:8] is selected here. (3) Obtain the product R-SMILES with specified root atom. (4) Remove the atom mapping to get the final input. (5) From the left to the right of the reactant SMILES, look for the atom mapping that appears on the product SMILES. Once found, the atom is selected as the root of the reactant. [C:1] and [Cl:8] are selected here. (6) Obtain the reactant R-SMILES without atom mapping to get the final output. (7) Tokenize the SMILES.}
\begin{tabular*}{\textwidth}{@{\extracolsep{\fill}}clll}
\hline
      & \multicolumn{2}{l}{Step} & Example(id 66, USPTO-50K dataset): reactants $>>$ products \\
\hline
(1)   & \multicolumn{2}{l}{Original data} & \multicolumn{1}{p{32.75em}}{Cl[C:1]([CH:2]=[CH2:3])=[O:4].[OH:5][CH2:6][C:7]([Cl:8])([Cl:9])[Cl:10]} \\
      & \multicolumn{2}{c}{} & $>>$[C:1]([CH:2]=[CH2:3])(=[O:4])[O:5][CH2:6][C:7]([Cl:8])([Cl:9])[Cl:10] \\
(2)   & \multicolumn{2}{l}{Randomly select a root} & [C:1]([CH:2]=[CH2:3])(=[O:4])[O:5][CH2:6][C:7](\textbf{[Cl:8]})([Cl:9])[Cl:10] \\
      & \multicolumn{2}{l}{atom from product} &  \\
(3)   & \multicolumn{2}{l}{Product R-SMILES} & \multicolumn{1}{p{32.75em}}{\textbf{[Cl:8]}[C:7]([Cl:9])([Cl:10])[C:6][O:5][C:1](=[O:4])[C:2]=[C:3]} \\
      & \multicolumn{2}{l}{with root atom mapping} &  \\
(4)   & \multicolumn{2}{l}{Atom-mapping removal} & \textbf{Cl}C(Cl)(Cl)COC(=O)C=C \\
(5)   & \multicolumn{2}{l}{Select reactant roots} & \multicolumn{1}{p{32.75em}}{Cl\textbf{[C:1]}([CH:2]=[CH2:3])=[O:4].[OH:5][CH2:6][C:7](\textbf{[Cl:8]})([Cl:9])[Cl:10]} \\
      & \multicolumn{2}{l}{according to product} &  \\
(6)   & \multicolumn{2}{l}{Reactant R-SMILES} & \textbf{Cl}C(Cl)(Cl)CO.\textbf{C}(=O)(Cl)C=C \\
      & \multicolumn{2}{l}{without atom mapping} &  \\
(7)   & \multicolumn{2}{l}{Tokenization} &  \\
      & \multicolumn{2}{l}{Source} & \textbf{Cl} C ( Cl ) ( Cl ) C O C ( = O ) C= C \\
      & \multicolumn{2}{l}{Target} & \textbf{Cl} C ( Cl ) ( Cl ) C O . \textbf{C} ( = O ) ( Cl ) C = C \\
\hline
\end{tabular*}%
\label{table:root_alignment}%
\end{table*}

\section{Methods}
To thoroughly evaluate the performance of the R-SMILES proposed for synthesis prediction, we implement our method on different synthesis tasks, including reactant-to-product, product-to-reactant, product-to-synthon, and synthon-to-reactant.
The first two can be classified as the template-free method and the other two as the semi-template method.
Template-free methods~\cite{liu2017retrosynthetic,karpov2019transformer,zheng2019predicting,lin2020automatic,tetko2020state,Seo_Song_Yang_Bae_Lee_Shin_Hwang_Yang_2021,kim2021valid,sacha2021molecule,sun2021towards} learn a direct mapping between products and reactants. 
Here for simplicity, the product is abbreviated as P and the reactant as R.
The direct transformation between products and reactants is denoted by P2R or R2P.
Semi-template methods~\cite{yan2020retroxpert,wang2021retroprime,shi2020graph,somnath2021learning} decompose retrosynthesis into two stages: 1) first identify intermediate molecules called synthons, and then 2) complete synthons into reactants. 
We use S to represent synthons, and P2S and S2R to represent the two stages, respectively. 
These four tasks are all formulated as end-to-end seq2seq problems and solved by the same model architecture to make comparisons with state-of-the-art~(SOTA) methods.
Many existing retrosynthesis work~\cite{dai2019retrosynthesis,chen2021deep,shi2020graph,somnath2021learning,sacha2021molecule} demonstrate their performances with the reaction type known for each product. Since the reaction type is not always available in real-world scenarios, all experiments in this work are carried out without this information.

\subsection{Datasets and data preprocessing} Experiments are conducted on USPTO-50K~\cite{schneider2016s},  USPTO-MIT~\cite{jin2017predicting} and  USPTO-FULL~\cite{dai2019retrosynthesis}, all of which are widely used as public benchmarking datasets for the synthesis prediction task. USPTO-50K is a high-quality dataset containing about 50,000 reactions with accurate atom mappings between products and reactants. USPTO-MIT contains about 400, 000 reactions as the training set, 30, 000 reactions as the validation set and 40, 000 reactions as the test set. USPTO-FULL is a much larger dataset for chemical reactions, consisting of about 1,000,000 reactions. For retrosynthesis prediction, reactions that contain multiple products are duplicated into multiple reactions to ensure that every reaction in data has only one product. Invalid data that contains no products or just a single ion as reactants are removed.

We use the same data split as previous researchers~\cite{coley2017computer, jin2017predicting,dai2019retrosynthesis} for all the datasets.
During the pretraining stage, depending on whether it is a forward or retrosynthesis prediction, products or reacatants in the training set of USPTO-FULL are used for self-supervised training, where molecules in the test set of USPTO-50K and USPTO-MIT are removed.

\subsection{Root-aligned SMILES} 
First of all, we follow Schwaller \textit{et al.}'s~\cite{schwaller2018found} regular expression to tokenize SMILES to meaningful tokens. To get R-SMILES, we have to find the common structures of the source and the target, which can be found by atom mapping or substructure matching algorithms~\cite{englert2015efficient}. In this work, we use atom mapping in the reactions to find the common structures. 

The root alignment operation is effortless in the P2R stage, where the input is only a single product. We can select a root atom from the product randomly first, and set it as the root atom to obtain the product SMILES. According to the new order of product tokens, we can find each corresponding root atom for reactants. We remove all atom mapping from the final input and output to avoid any information leak. An example of the root alignment is shown in Table~\ref{table:root_alignment}. In the S2R stage, we put the product and synthon SMILES together as input, separated by a special token that does not exist in the SMILES syntax. We choose to align reactants to synthons to minimize the difference between the input and the output since there is a one-to-one mapping between synthons and reactants. The product is aligned to the largest synthon (\textit{i.e.}, the synthon with the most atoms). Taking the reaction in Table~\ref{table:root_alignment} as the example, first we can get the synthon with atom-mapping that is ``[C:1]([CH:2]=[CH2:3])=[O:4].[O:5][CH2:6][C:7].
\noindent([Cl:8])([Cl:9])[Cl:10]''. By selecting [Cl:8] and [C:1] as the roots of the synthons, we can obtain the input as ``Cl C ( Cl ) ( Cl ) C O C ( = O ) C= C <split> Cl C ( Cl ) ( Cl ) C O . C ( = O ) C= C'' and the output as ``Cl C ( Cl ) ( Cl ) C O . C ( = O ) ( Cl ) C = C''.  In the R2P stage, we align the product SMILES to the largest reactant. After root alignment, the input and output are highly similar to each other, which helps the model to reduce the search space and makes cross-attention stronger.

\subsection{Data augmentation with R-SMILES}
Following the data augmentation strategy of the previous researchers~\cite{Seo_Song_Yang_Bae_Lee_Shin_Hwang_Yang_2021,wang2021retroprime,tetko2020state}, we apply $20\times$ augmentation at training and test sets of USPT0-50K, and $5\times$ augmentation at training and test sets of USPTO-MIT and USPTO-FULL.
When training the model, by enumerating different atoms as the root of SMILES, we can obtain multiple input-output pairs as the training data.
In the inference stage, we input several different SMILES representing the same input to obtain multiple sets of outputs. Then we acquire the final prediction result by scoring these outputs uniformly. You can find the detail of how to make model predictions with data augmentation in the supplementary information.

To highlight the superiority of R-SMILES, we use the vanilla transformer~\cite{vaswani2017attention} without any modification. The source code is available online at https://github.com/otori-bird/retrosynthesis. The detailed descriptions of the model architecture and training details are available in the supplementary information.

\begin{table}[htbp]
\caption{\ Edit distance with/without root alignment. Except for the data size, all figures are shown on average. Dataset$^{\times m}$: $m$  times data augmentation. Pro.: product SMILES. Rea.: reactant SMILES.}
\resizebox{0.48\textwidth}{!}{
\begin{tabular}{cccccc}
\hline
\multirow{2}[4]{*}{dataset} & \multirow{2}[4]{*}{data size} & \multicolumn{2}{c}{length} & \multicolumn{2}{c}{edit distance} \\
\cmidrule{3-6}      &       & Pro. & Rea. & w/o &  w/ \\
\hline
USPTO-50K$^{\times1}$ & 50,016 & 43.4  & 47.4  & 17.9  & 14.1(-21\%) \\
USPTO-50K$^{\times5}$ & 250,060 & 45.1  & 49.6  & 28.3  & 14.1(-50\%) \\
USPTO-50K$^{\times10}$ & 500,160 & 45.3  & 49.9  & 30.0  & 14.1(-53\%) \\
USPTO-50K$^{\times20}$ & 1,000,240 & 45.4  & 50.0  & 30.2  & 14.1(-53\%) \\
USPTO-MIT$^{\times1}$ & 482,132 & 40.6  & 46.1  & 17.0  & 13.5(-21\%) \\
USPTO-MIT$^{\times5}$ & 2,410,660 & 41.6  & 47.0  & 26.7  & 13.5(-49\%) \\
USPTO-Full$^{\times1}$ & 960,198 & 41.4  & 48.1  & 19.8  & 16.6(-16\%) \\
USPTO-Full$^{\times5}$ & 4,800,990 & 43.1  & 50.4  & 29.2  & 16.6(-43\%) \\
\hline
\end{tabular}%
}
\label{table:edit_distance}%
\end{table}

\section{Results and discussion}

\subsection{Statistical Analysis of the minimum edit distance with R-SMILES}

We first provide some statistical analysis of the minimum edit distance between the input and the output for retrosynthesis prediction with or without the proposed R-SMILES in Table~\ref{table:edit_distance}. The minimum edit distance between two strings is defined as the minimum number of editing operations (including insertion, deletion, and substitution) needed to transform one into the other. Here we adopt it to measure the discrepancy between input and output SMILES. Without R-SMILES, the average minimum edit distance between product and reactant SMILES is 17.9 on USPTO-50K, 17.0 on USPTO-MIT, and 19.8 on USPTO-FULL. However, with the proposed R-SMILES, the minimum edit distances become 14.1, 13.5, and 16.6, decreasing by 21\%, 21\%, and 16\%, respectively. Moreover, to alleviate the overfitting problem, data augmentation with randomized SMILES is critical and widely used in existing methods~\cite{schwaller2019molecular,wang2021retroprime,tetko2020state,Seo_Song_Yang_Bae_Lee_Shin_Hwang_Yang_2021}, but it would inevitably lead to a significant increase in the edit distance. 
For example, with 5$\times$ augmentation, the minimum edit distance is increased to 28.4 on USPTO-50K, which is more than two times of that of the proposed R-SMILES~(14.1), where the minimum edit distance of R-SMILES keeps unchanged with data augmentation. The larger discrepancy and one-to-many mapping of randomized SMILES make the learning problem more difficult, hindering the performance of synthesis prediction.

\begin{table}[h]
\small
\caption{\ Top-K accuracy of forward synthesis on the USPTO-MIT dataset. ``Separated'' and ``Mixed'' denote whether reagents are separated from reactants or not.}
\label{table:r2p_table}%
\resizebox{0.48\textwidth}{!}{
\begin{tabular}{ccccccc}
\hline
\multicolumn{7}{c}{USPTO-MIT top-K Accuracy (\%)} \\
\hline
\multicolumn{1}{l}{Setting} & \multicolumn{1}{l}{Model}  & K = 1 & 2     & 5     & 10    & 20 \\
\hline
\multirow{5}[2]{*}{Separated} 
& MT~\cite{schwaller2019molecular,sacha2021molecule}    & 90.5  & 93.7   & 95.3  & 96.0  & 96.5  \\
      & MEGAN~\cite{sacha2021molecule}   & 89.3  & 92.7  & 95.6  & 96.7  & 97.5  \\
      & AT~\cite{tetko2020state}     & 91.9  & 95.4   & 97.0  & -     & - \\
      & Chemformer~\cite{irwin2022chemformer}     & \textbf{92.8}  & -   & 94.9  & 95.0     & - \\
      & Ours   & 92.3  & \textbf{95.8 }  & \textbf{97.5 } & \textbf{98.0 } & \textbf{98.6 } \\
\hline
\multirow{5}[2]{*}{Mixed} & MT~\cite{schwaller2019molecular,sacha2021molecule}    & 88.7  & 92.1  & 94.2  & 94.9  & 95.4  \\
      & MEGAN~\cite{sacha2021molecule}   & 86.3  & 90.3    & 94.0  & 95.4  & 96.6  \\
      & AT~\cite{tetko2020state}    & 90.4  & 94.6   & 96.5  & -     & - \\
      & Chemformer~\cite{irwin2022chemformer}     & \textbf{91.3}  & -   & 93.7 & 94.0     & - \\
      & Ours  & 91.0  & \textbf{95.0 } & \textbf{96.8 } & \textbf{97.0 } & \textbf{97.3 } \\
\hline
\end{tabular}
}
\end{table}

\begin{table}[h]
\small
\caption{\ Top-K accuracy in P2S and S2R stages on the USPTO-50K dataset.}
\label{table:p2s_s2r}%
\resizebox{0.48\textwidth}{!}{
\begin{tabular}{llcccc}
\hline
\multicolumn{6}{c}{USPTO-50K top-K Accuracy(\%)} \\
\hline
Stage & Model & K=1   & 3     & 5     & 10 \\
\hline
\multirow{4}[1]{*}{P2S} & G2Gs~\cite{shi2020graph}  & 75.8  & 83.9  & 85.3  & 85.6  \\
  & GraphRetro~\cite{somnath2021learning} & 70.8  & 92.2  & 93.7  & 94.5  \\
  & RetroPrime~\cite{wang2021retroprime}  & 65.6  & 87.7  & 92.0  & - \\
  & Ours  & 75.2  & \textbf{94.4 } & \textbf{97.9 } & \textbf{99.1 } \\
\hline
\multirow{4}[2]{*}{S2R} & G2Gs~\cite{shi2020graph}  & 61.1  & 81.5  & 86.7  & 90.0  \\
  & GraphRetro~\cite{somnath2021learning}  & 75.6  & 87.7  & 92.9  & 96.3  \\
  & RetroPrime~\cite{wang2021retroprime}  & 73.4  & 87.9  & 89.8  & 90.4  \\
  & Ours  & 73.9  & \textbf{91.9 } & \textbf{95.2 } & \textbf{97.4 } \\
\hline
\end{tabular}
}
\end{table}

\subsection{Comparisons with SOTA methods}
We make comparisons between the proposed method and existing SOTA competitors for all four tasks.
Top-K exact match accuracy, which represents the percentage of predicted reactants that are identical to the ground truth, is adopted as the metric to evaluate the performance.
We additionally adopt the maximal fragment accuracy~\cite{tetko2020state} to evaluate the performance of P2R. The maximal fragment accuracy~(MaxFrag), inspired by classical retrosynthesis, requires the exact match of only the largest reactant. 
The top-K exact match accuracy is used as the main metric to report the performance, and the maximal fragment accuracy is adopted in some cases for a more comprehensive comparison. Experiments are conducted on USPTO-50K, USPTO-MIT, and USPTO-FULL datasets.

\begin{table*}[ht]
\small
\caption{\ Top-K single-step retrosynthesis results on USPTO-50K~(top), USPTO-MIT~(middle), and USPTO-FULL~(bottom) datasets}
\label{table:p2r_table}%

\resizebox{\textwidth}{!}{
\begin{tabular}{llcccccc}
\hline
\multicolumn{8}{c}{USPTO-50K top-K Accuracy (\%)} \\
\hline
\multicolumn{1}{l}{Category} & \multicolumn{1}{l}{Model} & K = 1 & 3     & 5     & 10    & 20    & 50 \\
\hline
\multirow{4}[2]{*}{Template-Based} & retrosim \cite{coley2017computer}  & 37.3  & 54.7  & 63.3  & 74.1  & 82.0  & 85.3  \\
  & neuralsym \cite{segler2017neural}   & 44.4  & 65.3   & 72.4  & 78.9  & 82.2  & 83.1  \\
  & GLN \cite{dai2019retrosynthesis}  & 52.5  & 69.0  & 75.6  & 83.7  & 89.0  & 92.4  \\
  & LocalRetro \cite{chen2021deep}  & 53.4  & 77.5  & 85.9  & 92.4  & -     & 97.7  \\
\hline
\multirow{5}[2]{*}{Semi-Template} & G2Gs \cite{shi2020graph}  & 48.9  & 67.6  & 72.5  & 75.5  & -     & - \\
  & GraphRetro \cite{somnath2021learning}  & 53.7  & 68.3  & 72.2  & 75.5  & -     & - \\
  & RetroXpert \cite{yan2020retroxpert}  & 50.4  & 61.1  & 62.3  & 63.4  & 63.9  & 64.0  \\
  & RetroPrime \cite{wang2021retroprime}   & 51.4  & 70.8  & 74.0  & 76.1  & -     & - \\
  & Ours\textsuperscript{\emph{b}}   & 49.1 ±0.42  & 68.4 ±0.53  & \textbf{75.8 ±0.62}  & \textbf{82.2 ±0.72}  & \textbf{85.1 ±0.81}     & \textbf{88.7 ±0.88} \\
\hline
\multirow{12}[2]{*}{Template-Free} 
  & Liu's Seq2seq \cite{liu2017retrosynthetic}   & 37.4  & 52.4  & 57.0  & 61.7  & 65.9  & 70.7  \\
  & Levenshtein \cite{sumner2020levenshtein}   & 41.5  & 48.1  & 50.0  & 51.4  & -  & -  \\
  & GTA \cite{Seo_Song_Yang_Bae_Lee_Shin_Hwang_Yang_2021}  & 51.1 ±0.29 & 67.6 ±0.22 & 74.8 ±0.36 & 81.6 ±0.22 & -     & - \\
  & Dual-TF \cite{sun2021towards} & 53.3  & 69.7  & 73.0  & 75.0  & -     & - \\
  & MEGAN \cite{sacha2021molecule}  & 48.1  & 70.7  & 78.4  & 86.1  & 90.3  & 93.2  \\
  & Tied Transformer \cite{kim2021valid}   & 47.1  & 67.2  & 73.5  & 78.5  & -     & - \\
  & AT \cite{tetko2020state}   & 53.5  & -     & 81.0  & 85.7  & -     & - \\
& Ours\textsuperscript{\emph{a}}   & \textbf{57.5 ±0.15} & \textbf{80.4 ±0.28} & \textbf{87.2 ±0.34} & \textbf{91.7 ±0.46} & \textbf{93.6 ±0.48} & \textbf{95.0 ±0.56} \\
  & MEGAN \cite{sacha2021molecule}~(MaxFrag)  & 54.2  & 75.7  & 83.1  & 89.2  & 92.7  & 95.1  \\
  & Tied Transformer \cite{kim2021valid}~(MaxFrag)  & 51.8  & 72.5  & 78.2  & 82.4  & -     & - \\
  & AT \cite{tetko2020state}~(MaxFrag) & 58.5  & -     & 85.4  & 90.0  & -     & - \\
    & Ours\textsuperscript{\emph{a}}~(MaxFrag)   & \textbf{61.0 ±0.14} & \textbf{82.5 ±0.26} & \textbf{88.5 ±0.30} & \textbf{92.8 ±0.35} & \textbf{94.6 ±0.45} & \textbf{95.7 ±0.53} \\
\hline
\multicolumn{8}{c}{USPTO-MIT top-K Accuracy (\%)} \\
\hline
\multicolumn{1}{l}{Category} & \multicolumn{1}{l}{Model}  & K = 1 & 3     & 5     & 10    & 20    & 50 \\
\hline
\multirow{2}[2]{*}{Template-Based}
  & neuralsym \cite{segler2017neural}  & 47.8  & 67.6     & 74.1     &  80.2 & -     & - \\
  & LocalRetro \cite{chen2021deep}  & 54.1 & 73.7     & 79.4     & 84.4  & -     & 90.4 \\
\hline
\multirow{4}[2]{*}{Template-Free} 
  & Liu's Seq2seq \cite{liu2017retrosynthetic}   & 46.9  & 61.6     & 66.3     & 70.8  & -     & - \\
  & AutoSynRoute \cite{lin2020automatic}    & 54.1  & 71.8     & 76.9     & 81.8  & -     & -  \\
  & RetroTRAE \cite{ucak2022retrosynthetic}   & 58.3  & -     & -     & -  & -     & - \\
  & Ours\textsuperscript{\emph{a}} & \textbf{60.3 ±0.22} & \textbf{78.2 ±0.28} & \textbf{83.2 ±0.36} & \textbf{87.3 ±0.38} &    \textbf{89.7 ±0.35}    & \textbf{91.6 ±0.44} \\
\hline
\multicolumn{8}{c}{USPTO-FULL top-K Accuracy (\%)} \\
\hline
\multicolumn{1}{l}{Category} & \multicolumn{1}{l}{Model}  & K = 1 & 3     & 5     & 10    & 20    & 50 \\
\hline
\multirow{4}[2]{*}{Template-Based} & retrosim \cite{coley2017computer} & 32.8  & -     & -     & 56.1  & -     & - \\
  & neuralsym \cite{segler2017neural} & 35.8  & -     & -     & 60.8  & -     & - \\
  & GLN \cite{dai2019retrosynthesis}  & 39.3  & -     & -     & 63.7  & -     & - \\
  & LocalRetro \cite{chen2021deep}\textsuperscript{\emph{c}}  & 39.1  & 53.3     & 58.4     & 63.7  & 67.5     & 70.7 \\
  
\hline
Semi-Template & RetroPrime \cite{wang2021retroprime}  & 44.1  & -     & -     & 68.5  & -     & - \\
\hline
\multirow{4}[2]{*}{Template-Free} & MEGAN \cite{sacha2021molecule}   & 33.6  & -     & -     & 63.9  & -     & 74.1  \\
  & GTA \cite{Seo_Song_Yang_Bae_Lee_Shin_Hwang_Yang_2021}   & 46.6 ±0.20 & -     & -     & 70.4 ±0.15 & -     & - \\
  & AT \cite{tetko2020state}  & 46.2  & -     & -     & 73.3  & -     & - \\
    & Ours\textsuperscript{\emph{a}}   & \textbf{49.8 ±0.18} & \textbf{67.3 ±0.24} & \textbf{72.5 ±0.34} & \textbf{77.5 ±0.40} &    \textbf{80.7 ±0.45}    & \textbf{83.4 ±0.52}
  \\
\hline
\end{tabular}
}
\textsuperscript{\emph{a}} Our product-to-reactant variant;
\textsuperscript{\emph{b}} Our product-to-synthon-to-reactant variant;
\textsuperscript{\emph{c}} Denotes that the result is implemented by the open-source code with well-tuned hyperparameters.
\end{table*}

Results of forward synthesis prediction are shown in Table~\ref{table:r2p_table}. Similar to Schwaller \textit{et al.}~\cite{schwaller2019molecular}, we conduct experiments in two settings: ``separated'' and ``mixed''. The latter is a more challenging task as the model has to recognize the reactants correctly. Except that MEGAN~\cite{sacha2021molecule} is a graph-based method, others are all transformer-based.
It is clear that our method outperforms others in most cases.
Although Chemformer~\cite{irwin2022chemformer} uses much more model parameters and data  than ours for pretraining, our method still obtains better results with the exception of  top-1 accuracy.
In different settings, the top-5 accuracy of our method is equal to or even higher than the top-20 accuracy of MEGAN, which fully illustrates the high efficiency of our method.

\begin{figure*}[ht!]
    \centering
    \includegraphics[width=\textwidth]{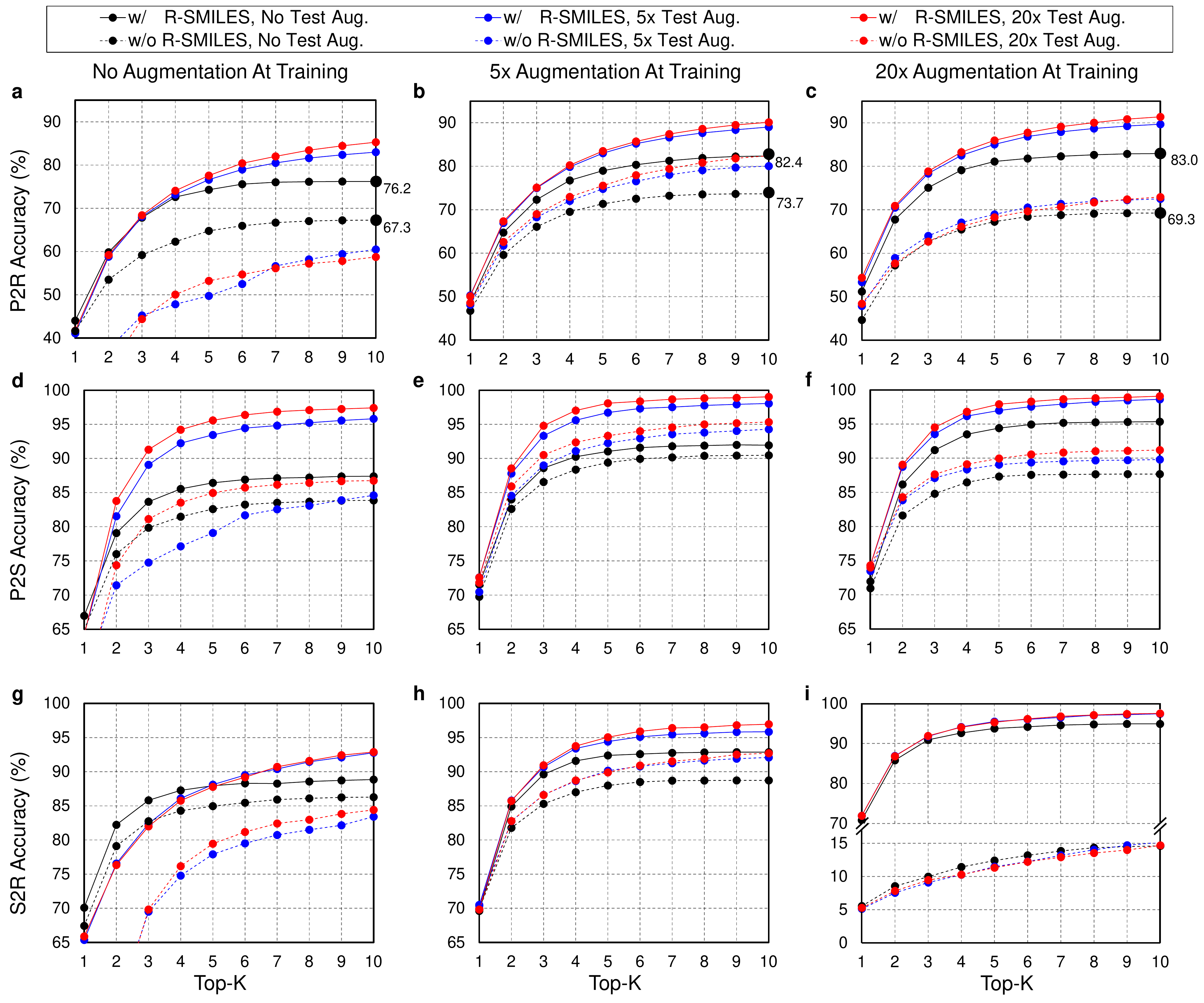}
    \caption{Top-K accuracy (\%) with/without R-SMILES on USPTO-50K for P2R~(a, b, c), P2S~(d, e, f), and S2R~(g, h, i).
    The solid lines (w/ R-SMILES) and dashed lines (w/o R-SMILES) represent the performance with or without R-SMILES, respectively.
    The lines with different colors represent the performance in different test set augmentation scenarios.
    }
    \label{fig:augmentation}
\end{figure*}

Results of retrosynthesis prediction are shown in Table~\ref{table:p2s_s2r} and Table~\ref{table:p2r_table}, from which we make the following three main conclusions:
1) Generally speaking, the proposed P2R variant consistently outperforms SOTA competitors by a large margin. On the USPTO-50K dataset, it outperforms the current best template-free method by absolute 4.0\%, 5.6\% and 1.8\% in top-1, top-10 and top-50 exact match accuracy, respectively. On the USPTO-MIT dataset, it also outperforms the concurrent work RetroTRAE~\cite{ucak2022retrosynthetic} that only reports the top-1 accuracy, and yields better performance at other top-K accuracies than any other method. On the more challenging USPTO-FULL dataset, the accuracy improvement is still very substantial, by 3.2\% in top-1, 4.2\% in top-10, and 9.3\% in top-50.
Similarly, our P2S and S2R variants also achieve the best results except for the top-1 accuracy on the USPTO-50K dataset. The top-10 accuracies of them even reach 99.1\% and 97.4\%, respectively.
We also combine these two phases together to get our product-to-synthon-to-reactant method that outperforms the current best semi-template method  by absolute 1.8\% and 6.1\% in top-5 and top-10 accuracy, respectively.
These impressing and consistent results demonstrate the superiority of the proposed method over SOTA methods.
2) Although Levenshtein augmentation~\cite{sumner2020levenshtein} ensures the high similarity between the input and output SMILES as we do, it cannot guarantee the one-to-one mapping between them, which largely inhibits its performance.
By specifying the root atom of input and output SMILES, our method can effectively guarantee the one-to-one mapping between them.
3) Our P2R variant achieves superior or at least comparable performance to the current SOTA template-based method LocalRetro~\cite{chen2021deep} on the USPTO-50K dataset. However, as template-based approaches are well known to be poor at generalizing to new reaction templates and coping with the huge number of reaction templates, the performances of LocalRetro on two large datasets USPTO-MIT and USPTO-FULL are substantially worse than ours, which strongly demonstrates the limitations of template-based methods.
All these results verify the effectiveness and the superiority of our proposed method.

\subsection{Superiority of the proposed R-SMILES with data augmentation}
Here we evaluate the superiority of the proposed R-SMILES when data augmentation is applied in retrosynthesis tasks. We adopt the vanilla transformer~\cite{vaswani2017attention}, a popular language translation model, as the retrosynthesis model. In retrosynthesis prediction, data augmentation can be applied to both the training and the test data~\cite{tetko2020state}, or only one of them. To test the performance of R-SMILES with data augmentation, different times of augmentation are conducted on training and test data. Here we take the widely used canonical SMILES as the baseline for comparisons. Experiments are conducted on the USPTO-50K dataset, with P2R, P2S, and S2R variants. Results are shown in Fig.~\ref{fig:augmentation}. In each subplot, the solid and dashed lines represent the performance with and without R-SMILES, and different colors represent times of data augmentation. First of all, it is evident that the solid lines are consistently above the dashed lines with the same color in each subplot, which reveals that the performance with R-SMILES is consistently superior to the widely used canonical SMILES in the same data augmentation scenario. An interesting observation is that if no training data augmentation is applied (Fig.~\ref{fig:augmentation}a, d, g), doing augmentation on the test data usually lowers the performance with the canonical SMILES. However, with the proposed R-SMILES, the accuracy is improved as expected, which indicates that the proposed method is more compatible with test data augmentation even though augmentation is not applied at the training time. Finally, by making plot-level comparisons, we can find that with more training data augmentation, the proposed R-SMILES yield higher accuracy. For example, if no data augmentation is applied at test time, 5$\times$ and 20$\times$ data augmentation of the training set increase the top-10 accuracy from 76.2\% to  82.4\% and 83.0\%, respectively. However, without R-SMILES, the model may yield inferior performance if too much training data augmentation is applied. In the same case as the example above, 5$\times$ data augmentation increases top-10 accuracy from 67.3\% to 73.7\%, but 20$\times$ augmentation decreases it to only 69.3\%. The underlying reason is that if too much training data augmentation is applied without R-SMILES, the retrosynthesis task becomes a one-to-many problem mentioned in Fig.~\ref{fig:R-SMILES}, which is extremely difficult for the model to learn useful chemical knowledge for retrosynthesis. However, if no training data augmentation is used, the model may easily suffer from the overfitting problem, which leaves a trade-off issue regarding the data augmentation. From the experimental results in Fig.~\ref{fig:augmentation}, it can be clearly seen that our proposed R-SMILES perfectly solves this issue and can reliably enjoy the higher performance with more data augmentation until reaching saturation.

\begin{figure}
    \centering
    \includegraphics[width=.5\textwidth]{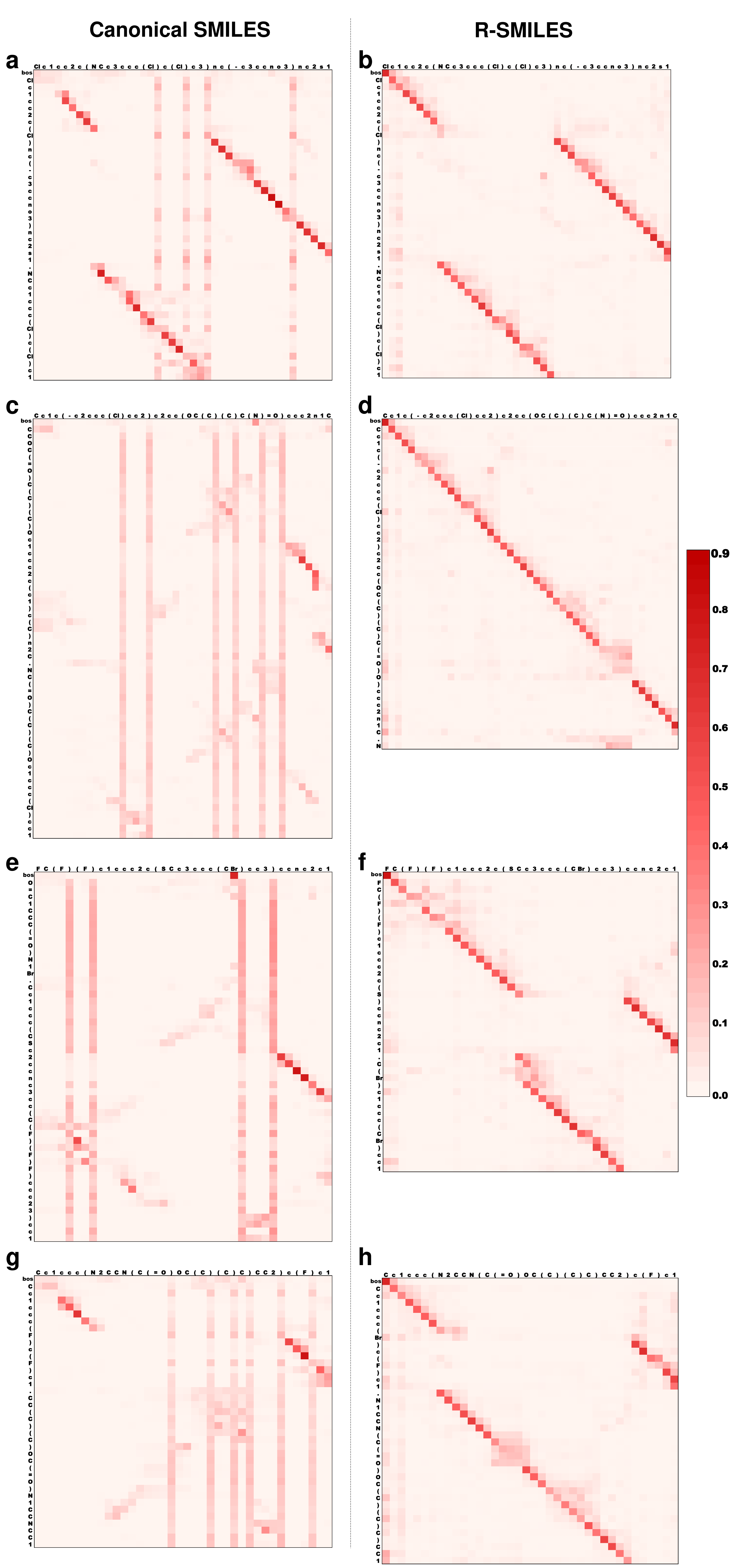}
    \caption{Visualization of the cross-attention obtained by the canonical SMILES~(Left) and the proposed R-SMILES~(Right) in the retrosynthesis prediction.
    (a, c, e, g) The attention maps obtained by the model trained with canonical SMILES.
    (b, d, f, h) The attention maps obtained by the model trained with R-SMILES.
     The input tokens are along the x axis, and the output tokens are along the y axis.
     Each row in the attention map represents the attention over the input tokens for predicting the next output token.
     Each column represents the attention between an input token with each output token.
     The ``bos'' token is the beginning of output tokens and will be removed after the decoding process completes.
    }
    \label{fig:attention}
\end{figure}

\subsection{Visualization of cross-attention mechanism in transformer with R-SMILES}
To further illustrate how the transformer works with R-SMILES, we randomly selected four reactions and display the visualization of the cross-attention maps in the retrosynthesis prediction in Fig.~\ref{fig:attention}.
The adopted transformer is an autoregressive model, where the last predicted token is taken as input for predicting the next token. The cross-attention represents the correlation between reactant tokens and product tokens.
By feeding the same canonical SMILES to the models trained with R-SMILES or canonical SMILES and averaging the attention of each attention head in the last layer of the Transformer Decoder, we can get these attention maps to make a direct comparison. 
In Fig.~\ref{fig:attention}a where the canonical SMILES of the product and the target reactant is highly similar to each other, it can be seen that the model could capture the aligned tokens and made the correct predictions. However, the attention of output tokens tended to pay much attention to some input tokens related to the SMILES syntax like `)', and this problem exists in all maps obtained by the model trained with canonical SMILES.
In contrast, with the proposed R-SMILES, the model gave the attention in Fig.~\ref{fig:attention}b that is paid more on corresponding tokens and also succeeded.
In Fig.~\ref{fig:attention}c, although the canonical SMILES of the product and the target reactant is also highly similar, the model gave a disordered attention map and failed, which indicates that its ability to capture alignment information is insufficient. 
However, the model trained with R-SMILES not only obtained a well-aligned attention map in Fig.~\ref{fig:attention}d, but also correctly predicted the target R-SMILES, where the target R-SMILES is also the canonical SMILES.
In Fig.~\ref{fig:attention}e, g where the canonical SMILES of the product and the target reactant is quite different, the model trained with canonical SMILES was unable to find alignment and had to focus on the global information, which ultimately led to the disordered attention maps and the failure of the predictions.
However, thanks to the small discrepancy of R-SMILES pairs, in Fig.~\ref{fig:attention}f, h the model trained with R-SMILES gave ordered attention maps and succeeded to predict the target R-SMILES.
These results all demonstrate that our proposed R-SMILES effectively allows the model to focus on learning chemical knowledge for reactions and thus improves the accuracy of the model prediction.
The attention maps of the forward reaction prediction and other layers can be found Fig. S3 and Fig. S4, from which  the same conclusion can be drawn.

\begin{figure*}
    \centering
    \includegraphics[width=1.00\textwidth]{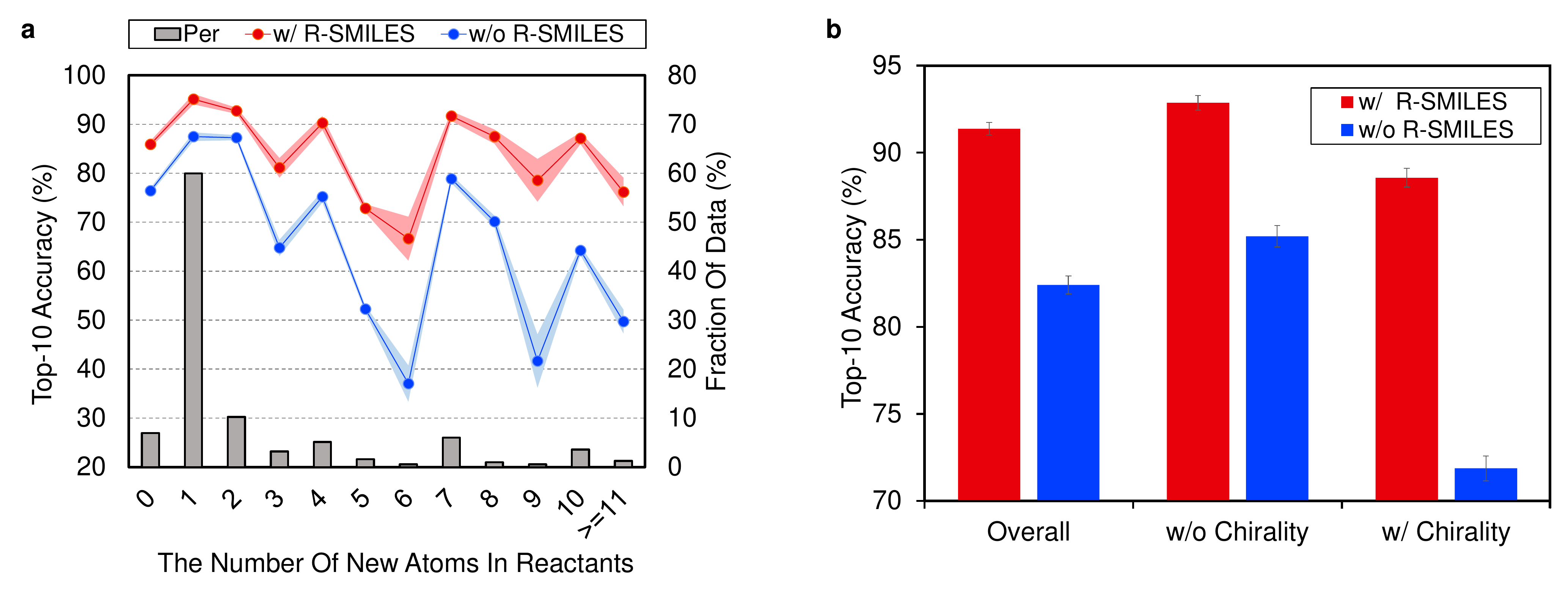}
       \caption{Accuracies for complex reactions.
    (a) Top-10 accuracy according to the number of new atoms in reactants.
    The red and blue lines represent the performance with/without R-SMILES.
    The gray bar means the percentage of this kind of reaction in the test set.
    (b) Top-10 accuracy for reactions involving with/without chirality.
    The red and blue bars represent the performance with or without R-SMILES.}
    \label{fig:accuracy_bymulti}
\end{figure*}

\subsection{Evaluating R-SMILES in more aspects of retrosynthesis}
Here we conduct further studies to shed more light on the proposed R-SMILES when applied to retrosynthesis. Specifically, we investigate the performance of R-SMILES with some more complex reactions in the USPTO-50K, including reactions involving many new atoms in the reactants and chirality.

\subsubsection{The number of new atoms in reactants} 
According to the number of new atoms (hydrogen atoms do not count) in reactants, we illustrate top-10 accuracy with or without R-SMILES and the amount of data in Fig.~\ref{fig:accuracy_bymulti}a. Similar to the previous results, the red line is always above the blue line, illustrating that the performance with R-SMILES surpasses the other by a large margin. In addition, the more new atoms in reactants, the larger improvement, especially for the situations with small amounts of data. For the reactions whose numbers of new atoms are 9, the improvement is impressively 39.3\%, demonstrating that R-SMILES remains robust even with small amounts of data. This is because with R-SMILES that reduces the differences between the input and the output SMILES, the model can pay attention to the new fragments in the output SMILES.

\subsubsection{Chirality}
Chirality is a property of asymmetry and is important in drug discovery and stereochemistry. It can be represented by `@' or `@@' in SMILES sequences. We count 935 reactions with chirality in our test set of USPTO-50K and exhibit the top-10 accuracy with or without chirality and overall accuracy in Fig.~\ref{fig:accuracy_bymulti}b.
When chirality exists in the reaction, the accuracy without R-SMILES drops 13.3\%. 
In comparison, ours drops only 4.3\%, proving that even in the presence of chirality, R-SMILES can still help the model focus on the more meaningful differences between the input and output SMILES.
To be more specific,
we believe that R-SMILES helps the chiral reaction mainly in two ways: 1) As shown in Table S1, the reduction of editing distance of the chiral reaction is more significant than the overall one. 2)For USPTO datasets, the chiral signatures of the input and output tend to be identical after alignment, which makes the model usually only need to maintain the chiral consistency.

For other top-K accuracies, results for both indicators are similar and can be found in Figure S5. 
These results all demonstrate the effectiveness and robustness of R-SMILES.

\begin{figure*}
    \centering
    \includegraphics[width=1.00\textwidth]{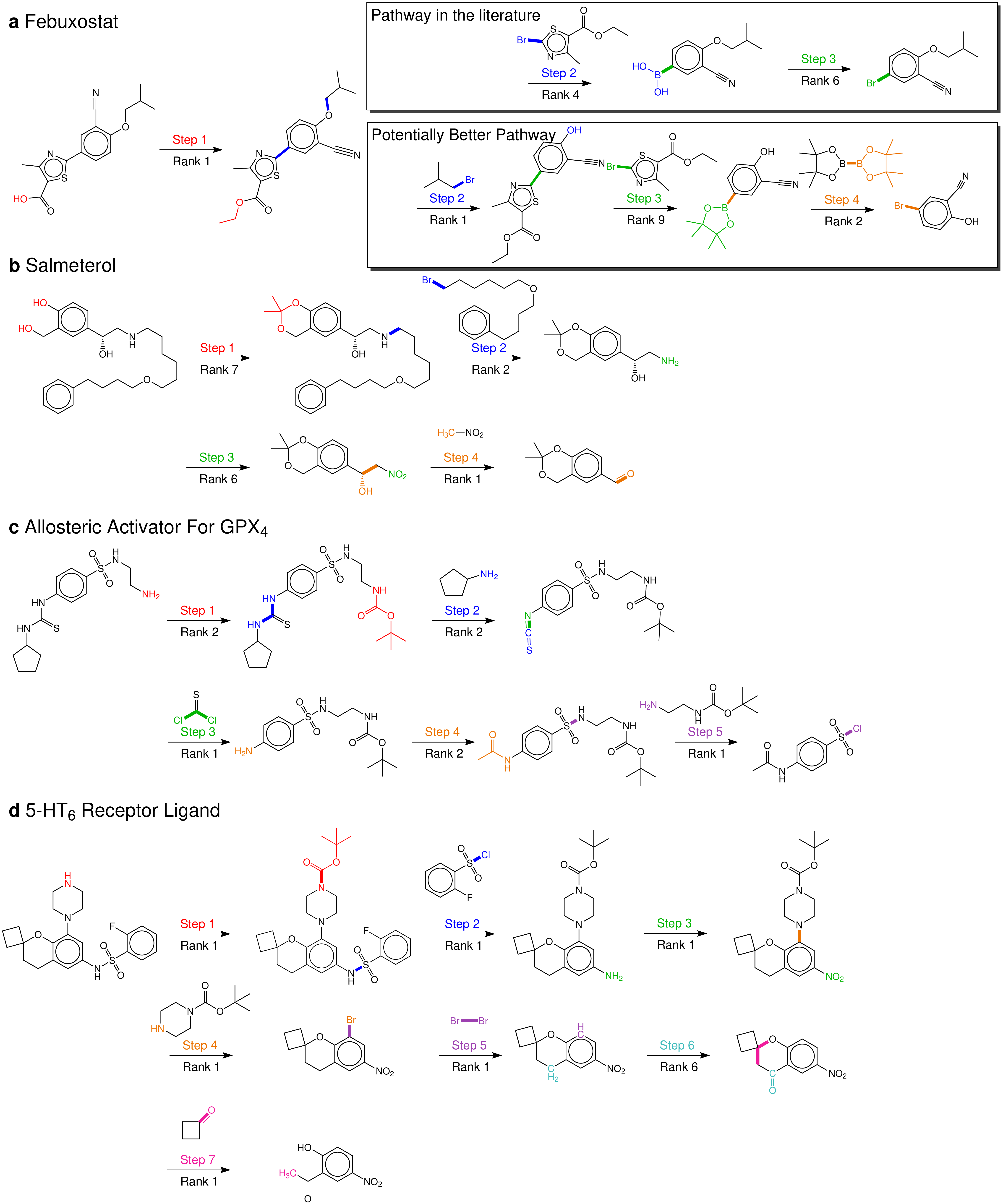}
    \caption{Multistep retrosynthesis predictions by our method.
    (a) Febuxostat.
    (b) Salmeterol. 
    (c) An allosteric activator for GPX$_4$. 
    (d) A 5-HT$_6$ receptor ligand.
    The reaction centers and transformations from products to reactants are highlighted in different colors at different reaction steps.
    In addition to the reaction pathway in the literature, we report a potentially better reaction pathway for febuxostat.
    }
    \label{fig:multistep_retosynthesis}
\end{figure*}

\subsection{Multistep retrosynthesis prediction by our method.}
By applying our product-reactant variant recursively, we verify our method with several multistep retrosynthesis examples reported in the literature, including febuxostat~\cite{2016The}, salmeterol~\cite{guo2011enantioselective}, an allosteric activator for GPX$_4$~\cite{lin2020automatic}, and a 5-HT$_6$ receptor ligand~\cite{nirogi2015design}. As shown in Fig.~\ref{fig:multistep_retosynthesis}, our method successfully predicts the complete synthetic pathway for these examples.

Febuxostat (Fig.~\ref{fig:multistep_retosynthesis}a) is a novel anti-gout drug as the non-purine selective inhibitor of xanthine oxidase. Cao \textit{et al.}~\cite{2016The} reported a new reaction pathway for it based on the Suzuki cross-coupling reaction in 2016. Our predicted first step is hydrolysis of the ester, which is exactly the same as reported. For the remaining reaction steps, our method provides two different synthetic routes. The first one is the same as reported, where 3-cyano-4-isobutoxyphenyl boronic acid and ethyl 2-bromo-4-methyithiazole-5-carboxylate are taken as the reactants of the Suzuki cross-coupling reaction. However, the second one reports nucleophilic substitution to get aryl boronic esters for the Suzuki cross-coupling reaction. 
The final steps of them both involve borylation, where the second one is reported by Ishiyama \textit{et al.}\cite{ishiyama1995palladium}. 
We can make a detailed comparison between these two pathways in terms of yield and price:
1) There are two main findings for us in Urawa et al.'s study~\cite{urawa2002investigations}: (a) Boronic acid is thermally less stable than the corresponding boronic ester. Thus, boronic ester is more likely to be better for avoiding possible thermal decomposition. (b) The introduction of pinacol boronate can effectively reduce the generation of side reactions, i.e., reductive dehalogenation reactions, which helps to afford the desired product quantitatively. The second synthetic pathway is consistent with these findings, which shows that the second is likely to have higher yields.
2) From the Reaxys database, it can be found that the building block of the second pathway is much cheaper compared to the first path.
Therefore, we believe that our method suggests a potentially better synthetic pathway for febuxostat.

Salmeterol (Fig.~\ref{fig:multistep_retosynthesis}b) is a potent, long-acting, $\beta$2-adrenoreceptor agonist. 
Guo \textit{et al.}~\cite{guo2011enantioselective} proposed a reaction pathway for it based on the asymmetric Henry reaction.
Although the first three steps provided by our method do not exist in the literature, they are all explainable. The first step reports the hydrolysis of cyclic acetal, where cyclic acetal has been proved to be stable. 
Considering the high activity of the phenolic hydroxyl group and the hydroxyl group connected to the benzyl group, the formation of cyclic acetal can effectively prevent the occurrence of side reactions, which illustrates the model has distinguished the properties of protection groups and preserved them to the starting compound. 
The second step involves the amination of halohydrocarbon, and the third step involves the reduction of the nitro group. The final step, which is the core reaction, is the asymmetric Henry reaction, where our method has successfully reproduced the generation of new chiral centers at the rank-1 prediction. 
This result also matches our conclusion of the great performance involving chirality as mentioned above.

The synthetic pathway of the GPX$_4$ activator compound (Fig.~\ref{fig:multistep_retosynthesis}c) is reported by Lin \textit{et al.}~\cite{lin2020automatic}, who predicted the synthetic pathway with a template-free model by enumerating different reaction types. However, even without the reaction type, our method succeeds for all five reaction steps within the top-2 predictions, which directly demonstrates the superiority of our method. Among these five reaction steps, the Hinsberg reaction of the final step is the core reaction of the whole synthetic pathway. Our method succeeds in finding it at the rank-1 prediction.

Nirogi \textit{et al.}~\cite{nirogi2015design} proposed a benzopyran sulfonamide derivative as an antagonist of 5-HT$_6$ receptor (Fig.~\ref{fig:multistep_retosynthesis}d) in 2015. Although the synthetic pathway consists of seven reaction steps, our method succeeds at the rank-1 prediction for all steps except the sixth one predicted at rank-6. The second and fourth steps have attracted our attention, which are the Hinsberg reaction and Nucleophilic Aromatic Substitution reaction (SNAr). In the Hinsberg reaction, primary amines are able to react with benzenesulfonyl chloride. In SNAr, the meta-nitro group reduces the density of electron cloud, which is conducive to the occurrence of reaction. The success of key steps in the long synthetic pathway further demonstrates the robustness of our method.

For all 22 reactions in these four examples, our method succeeds at the top-10 predictions, and mostly at the top-2 predictions. In addition, our method proposes a novel synthetic pathway for febuxostat that is more consistent with experimental experience. These exciting results all demonstrate the great potential of our method for multistep retrosynthesis.

% Table generated by Excel2LaTeX from sheet 'Sheet10'
\begin{table}[htbp]
\centering
\caption{The edit distance and top-K accuracy of single-step retrosynthesis for ring and non-ring reactions on the USPTO-50K dataset.}
\resizebox{0.48\textwidth}{!}{
\begin{tabular}{cccccc}
\hline
\textbf{Reaction Type} & \textbf{edit distance} & \textbf{K=1} & \textbf{3} & \textbf{5} & \textbf{10} \\
\hline
Overall\textsuperscript{\emph{a}} & 30.2  & 49.9  & 68.5  & 75.0  & 80.2  \\
Non-ring reaction\textsuperscript{\emph{a}} & 29.7  & 53.0  & 71.5  & 78.0  & 83.3  \\
Ring-opening reaction\textsuperscript{\emph{a}} & 37.8  & 23.3  & 42.0  & 49.7  & 54.6  \\
Ring-forming reaction\textsuperscript{\emph{a}} & 27.6  & 26.4  & 37.1  & 40.0  & 45.0  \\
\hline
Overall\textsuperscript{\emph{b}} & 14.1 (-53\%) & 56.3  & 79.1  & 86.0  & 91.0  \\
Non-ring reaction\textsuperscript{\emph{b}} & 13.3 (-55\%) & 58.8  & 81.5  & 88.5  & 93.1  \\
Ring-opening reaction\textsuperscript{\emph{b}} & 23.4 (-38\%) & 30.7  & 56.3  & 61.9  & 65.9  \\
Ring-forming reaction\textsuperscript{\emph{b}} & 17.5 (-37\%) & 38.0  & 57.9  & 63.6  & 71.9  \\
\hline
\end{tabular}%
}

\textsuperscript{\emph{a}} Without root alignment;
\textsuperscript{\emph{b}} With root alignment.
  \label{tab:ring_ed_acc}%
\end{table}%

\subsection{Limitations}
Even though our method currently achieves SOTA results on the USPTO datasets, the proposed R-SMILES has its own limitations. We calculated the accuracy of retrosynthesis for ring-opening and forming reactions in
different datasets. Results are shown in Table~\ref{tab:ring_ed_acc}. It can be seen that the accuracy of R-SMILES is not so high as that of other reactions. To make it clearer, we also calculated the edit distance between the input and the output SMILES for these reactions, as shown in Table~\ref{tab:ring_ed_acc}. Compared with that of non-ring reaction R-SMILES, the edit distance of ring reactions is significantly larger. These results again verify our main motivation in this work that large distance between input and output strings will degrade the reaction prediction performance. You can check the results of other datasets in Table S3.

The atom mapping annotations in the dataset may also be a limitation of the proposed method. Fortunately, in practice several fully automated atomic mapping tools have been developed, such as Indigo and RXNMapper~\cite{schwaller2021extraction}, which could be utilized for automatically generating the atom-mapping information. Albeit not perfectly accurate, these tools make the proposed method feasible on datasets without atom-mapping annotations. In fact, for the reported results on the USPTO-FULL dataset in our manuscript, all the R-SMILES are generated with the Indigo toolkit. The proposed method, as shown in Table 5, outperforms other competitors at any top-k accuracy. We believe these results give us a glimpse at the effectiveness of the proposed method on datasets without any atom-mapping annotations.

\section{Conclusions}
 
In this article, we propose R-SMILES for chemical reaction prediction. Unlike canonical SMILES that is widely adopted in the current literature, R-SMILES specifies a tightly aligned one-to-one mapping between the input and output SMILES, which decreases the edit distance significantly. With R-SMILES, the synthesis prediction model is largely relaxed from learning the complex syntax and can be dedicated to learning the chemical knowledge for reactions. We implement different variants to validate the proposed R-SMILES, both yielding superior performance to state-of-the-art methods. To better understand the proposed method, we further provide several interesting discussions, \textit{e.g.} the visualization of the cross-attention between input and output tokens. Finally, the synthetic pathways of some organic compounds are successfully predicted to showcase the effectiveness of the proposed method.

Albeit striking performance achieved in retrosynthesis, we believe that the potential of R-SMILES is not fully explored in this work. From the perspective of methods, since R-SMILES maintains the high similarity of the input and the output, retrosynthesis can be formulated as a grammatical error correction problem rather than a translation from scratch. 
To address the limitations mentioned above, in the future we will also try to align multiple atoms to obtain more similar input-output pairs, as well as to combine our method with the latest automatic atom mapping method for the datasets without atom mapping annotations.

\bibliography{sn-article}% common bib file

\begin{thebibliography}{10}
\expandafter\ifx\csname url\endcsname\relax
  \def\url#1{\burl{#1}}\fi
\expandafter\ifx\csname urlprefix\endcsname\relax\def\urlprefix{Preprint at
  }\fi
\providecommand{\bibinfo}[2]{#2}
\providecommand{\eprint}[2][]{\url{#2}}
\providecommand{\doi}[1]{\url{https://doi.org/#1}}
\bibcommenthead

\bibitem{pensak1977lhasa}
\bibinfo{author}{Pensak, D.~A.} \& \bibinfo{author}{Corey, E.~J.}
\newblock \bibinfo{title}{Lhasa—logic and heuristics applied to synthetic
  analysis}.
\newblock \emph{\bibinfo{journal}{ACS Symp. Ser.}}
  \textbf{\bibinfo{volume}{61}}, \bibinfo{pages}{1--32} (\bibinfo{year}{1977}).

\bibitem{johnson1989designing}
\bibinfo{author}{Johnson, P.} \emph{et~al.}
\newblock \bibinfo{title}{Designing an expert system for organic synthesis in
  expert systems application in chemistry}.
\newblock \emph{\bibinfo{journal}{ACS Symp. Ser.}}
  \textbf{\bibinfo{volume}{408}}, \bibinfo{pages}{102--123}
  (\bibinfo{year}{1989}).

\bibitem{gasteiger2000computer}
\bibinfo{author}{Gasteiger, J.} \emph{et~al.}
\newblock \bibinfo{title}{Computer-assisted synthesis and reaction planning in
  combinatorial chemistry}.
\newblock \emph{\bibinfo{journal}{Perspect. Drug Discovery Des.}}
  \textbf{\bibinfo{volume}{20}}~(1), \bibinfo{pages}{245--264}
  (\bibinfo{year}{2000}).

\bibitem{szymkuc2016computer}
\bibinfo{author}{Szymku{\'c}, S.} \emph{et~al.}
\newblock \bibinfo{title}{Computer-assisted synthetic planning: the end of the
  beginning}.
\newblock \emph{\bibinfo{journal}{Angew. Chem., Int. Ed.}}
  \textbf{\bibinfo{volume}{55}}~(20), \bibinfo{pages}{5904--5937}
  (\bibinfo{year}{2016}).

\bibitem{coley2017computer}
\bibinfo{author}{Coley, C.~W.}, \bibinfo{author}{Rogers, L.},
  \bibinfo{author}{Green, W.~H.} \& \bibinfo{author}{Jensen, K.~F.}
\newblock \bibinfo{title}{Computer-assisted retrosynthesis based on molecular
  similarity}.
\newblock \emph{\bibinfo{journal}{ACS Cent. Sci.}}
  \textbf{\bibinfo{volume}{3}}~(12), \bibinfo{pages}{1237--1245}
  (\bibinfo{year}{2017}).

\bibitem{segler2017neural}
\bibinfo{author}{Segler, M.~H.} \& \bibinfo{author}{Waller, M.~P.}
\newblock \bibinfo{title}{Neural-symbolic machine learning for retrosynthesis
  and reaction prediction}.
\newblock \emph{\bibinfo{journal}{Chem. - Eur. J.}}
  \textbf{\bibinfo{volume}{23}}~(25), \bibinfo{pages}{5966--5971}
  (\bibinfo{year}{2017}).

\bibitem{dai2019retrosynthesis}
\bibinfo{author}{Dai, H.}, \bibinfo{author}{Li, C.}, \bibinfo{author}{Coley,
  C.}, \bibinfo{author}{Dai, B.} \& \bibinfo{author}{Song, L.}
\newblock \bibinfo{title}{Retrosynthesis prediction with conditional graph
  logic network}.
\newblock In \emph{\bibinfo{booktitle}{Advances in Neural Information
  Processing Systems}}  (\bibinfo{publisher}{Curran Associates, Inc.},
  \bibinfo{year}{2019}).

\bibitem{chen2021deep}
\bibinfo{author}{Chen, S.} \& \bibinfo{author}{Jung, Y.}
\newblock \bibinfo{title}{Deep retrosynthetic reaction prediction using local
  reactivity and global attention}.
\newblock \emph{\bibinfo{journal}{JACS Au}} \textbf{\bibinfo{volume}{1}}~(10),
  \bibinfo{pages}{1612--1620} (\bibinfo{year}{2021}).

\bibitem{guo2020bayesian}
\bibinfo{author}{Guo, Z.}, \bibinfo{author}{Wu, S.}, \bibinfo{author}{Ohno, M.}
  \& \bibinfo{author}{Yoshida, R.}
\newblock \bibinfo{title}{Bayesian algorithm for retrosynthesis}.
\newblock \emph{\bibinfo{journal}{J. Chem. Inf. Model.}}
  \textbf{\bibinfo{volume}{60}}~(10), \bibinfo{pages}{4474--4486}
  (\bibinfo{year}{2020}).

\bibitem{lee2021retcl}
\bibinfo{author}{Lee, H.} \emph{et~al.}
\newblock \bibinfo{title}{Retcl: A selection-based approach for retrosynthesis
  via contrastive learning}.
\newblock In \emph{\bibinfo{booktitle}{Proceedings of the 31th International
  Joint Conference on Artificial Intelligence}} \bibinfo{pages}{2673--2679}
  (\bibinfo{publisher}{International Joint Conferences on Artificial
  Intelligence Organization}, \bibinfo{year}{2021}).

\bibitem{liu2017retrosynthetic}
\bibinfo{author}{Liu, B.} \emph{et~al.}
\newblock \bibinfo{title}{Retrosynthetic reaction prediction using neural
  sequence-to-sequence models}.
\newblock \emph{\bibinfo{journal}{ACS Cent. Sci.}}
  \textbf{\bibinfo{volume}{3}}~(10), \bibinfo{pages}{1103--1113}
  (\bibinfo{year}{2017}).

\bibitem{karpov2019transformer}
\bibinfo{author}{Karpov, P.}, \bibinfo{author}{Godin, G.} \&
  \bibinfo{author}{Tetko, I.~V.}
\newblock \bibinfo{title}{A transformer model for retrosynthesis}.
\newblock In \emph{\bibinfo{booktitle}{Artificial Neural Networks and Machine
  Learning -- ICANN: Workshop and Special Sessions}} \bibinfo{pages}{817--830}
  (\bibinfo{organization}{Springer}, \bibinfo{year}{2019}).

\bibitem{zheng2019predicting}
\bibinfo{author}{Zheng, S.}, \bibinfo{author}{Rao, J.}, \bibinfo{author}{Zhang,
  Z.}, \bibinfo{author}{Xu, J.} \& \bibinfo{author}{Yang, Y.}
\newblock \bibinfo{title}{Predicting retrosynthetic reactions using
  self-corrected transformer neural networks}.
\newblock \emph{\bibinfo{journal}{J. Chem. Inf. Model.}}
  \textbf{\bibinfo{volume}{60}}~(1), \bibinfo{pages}{47--55}
  (\bibinfo{year}{2019}).

\bibitem{lin2020automatic}
\bibinfo{author}{Lin, K.}, \bibinfo{author}{Xu, Y.}, \bibinfo{author}{Pei, J.}
  \& \bibinfo{author}{Lai, L.}
\newblock \bibinfo{title}{Automatic retrosynthetic route planning using
  template-free models}.
\newblock \emph{\bibinfo{journal}{Chem. Sci.}}
  \textbf{\bibinfo{volume}{11}}~(12), \bibinfo{pages}{3355--3364}
  (\bibinfo{year}{2020}).

\bibitem{yan2020retroxpert}
\bibinfo{author}{Yan, C.} \emph{et~al.}
\newblock \bibinfo{title}{Retroxpert: Decompose retrosynthesis prediction like
  a chemist}.
\newblock In \emph{\bibinfo{booktitle}{Advances in Neural Information
  Processing Systems}} \bibinfo{pages}{11248--11258}
  (\bibinfo{publisher}{Curran Associates, Inc.}, \bibinfo{year}{2020}).

\bibitem{wang2021retroprime}
\bibinfo{author}{Wang, X.} \emph{et~al.}
\newblock \bibinfo{title}{Retroprime: A diverse, plausible and
  transformer-based method for single-step retrosynthesis predictions}.
\newblock \emph{\bibinfo{journal}{Chem. Eng. J.}}
  \textbf{\bibinfo{volume}{420}}, \bibinfo{pages}{129845}
  (\bibinfo{year}{2021}).

\bibitem{tetko2020state}
\bibinfo{author}{Tetko, I.~V.}, \bibinfo{author}{Karpov, P.},
  \bibinfo{author}{Van~Deursen, R.} \& \bibinfo{author}{Godin, G.}
\newblock \bibinfo{title}{State-of-the-art augmented nlp transformer models for
  direct and single-step retrosynthesis}.
\newblock \emph{\bibinfo{journal}{Nat. Commun.}}
  \textbf{\bibinfo{volume}{11}}~(1), \bibinfo{pages}{1--11}
  (\bibinfo{year}{2020}).

\bibitem{Seo_Song_Yang_Bae_Lee_Shin_Hwang_Yang_2021}
\bibinfo{author}{Seo, S.-W.} \emph{et~al.}
\newblock \bibinfo{title}{Gta: Graph truncated attention for retrosynthesis}.
\newblock In \emph{\bibinfo{booktitle}{Proceedings of the AAAI Conference on
  Artificial Intelligence}} \bibinfo{pages}{531--539} (\bibinfo{year}{2021}).

\bibitem{kim2021valid}
\bibinfo{author}{Kim, E.}, \bibinfo{author}{Lee, D.}, \bibinfo{author}{Kwon,
  Y.}, \bibinfo{author}{Park, M.~S.} \& \bibinfo{author}{Choi, Y.-S.}
\newblock \bibinfo{title}{Valid, plausible, and diverse retrosynthesis using
  tied two-way transformers with latent variables}.
\newblock \emph{\bibinfo{journal}{J. Chem. Inf. Model.}}
  \textbf{\bibinfo{volume}{61}}~(1), \bibinfo{pages}{123--133}
  (\bibinfo{year}{2021}).

\bibitem{shi2020graph}
\bibinfo{author}{Shi, C.}, \bibinfo{author}{Xu, M.}, \bibinfo{author}{Guo, H.},
  \bibinfo{author}{Zhang, M.} \& \bibinfo{author}{Tang, J.}
\newblock \bibinfo{title}{A graph to graphs framework for retrosynthesis
  prediction}.
\newblock In \emph{\bibinfo{booktitle}{Proceedings of the 37th International
  Conference on Machine Learning}} \bibinfo{pages}{8818--8827}
  (\bibinfo{publisher}{PMLR}, \bibinfo{year}{2020}).

\bibitem{somnath2021learning}
\bibinfo{author}{Somnath, V.~R.}, \bibinfo{author}{Bunne, C.},
  \bibinfo{author}{Coley, C.}, \bibinfo{author}{Krause, A.} \&
  \bibinfo{author}{Barzilay, R.}
\newblock \bibinfo{title}{Learning graph models for retrosynthesis prediction}.
\newblock In \emph{\bibinfo{booktitle}{Advances in Neural Information
  Processing Systems}} \bibinfo{pages}{9405--9415} (\bibinfo{publisher}{Curran
  Associates, Inc.}, \bibinfo{year}{2021}).

\bibitem{sacha2021molecule}
\bibinfo{author}{Sacha, M.} \emph{et~al.}
\newblock \bibinfo{title}{Molecule edit graph attention network: modeling
  chemical reactions as sequences of graph edits}.
\newblock \emph{\bibinfo{journal}{J. Chem. Inf. Model.}}
  \textbf{\bibinfo{volume}{61}}~(7), \bibinfo{pages}{3273--3284}
  (\bibinfo{year}{2021}).

\bibitem{schlichtkrull2018modeling}
\bibinfo{author}{Schlichtkrull, M.} \emph{et~al.}
\newblock \bibinfo{title}{Modeling relational data with graph convolutional
  networks}.
\newblock In \emph{\bibinfo{booktitle}{The Semantic Web}}
  \bibinfo{pages}{593--607} (\bibinfo{organization}{Springer},
  \bibinfo{year}{2018}).

\bibitem{velickovic2017graph}
\bibinfo{author}{Velickovic, P.} \emph{et~al.}
\newblock \bibinfo{title}{Graph attention networks} (\bibinfo{year}{2017}).

\bibitem{weininger1988smiles}
\bibinfo{author}{Weininger, D.}
\newblock \bibinfo{title}{Smiles, a chemical language and information system.
  1. introduction to methodology and encoding rules}.
\newblock \emph{\bibinfo{journal}{J. Chem. Inf. Comput. Sci.}}
  \textbf{\bibinfo{volume}{28}}~(1), \bibinfo{pages}{31--36}
  (\bibinfo{year}{1988}).

\bibitem{nam2016linking}
\bibinfo{author}{Nam, J.} \& \bibinfo{author}{Kim, J.}
\newblock \bibinfo{title}{Linking the neural machine translation and the
  prediction of organic chemistry reactions} (\bibinfo{year}{2016}).

\bibitem{schwaller2018found}
\bibinfo{author}{Schwaller, P.}, \bibinfo{author}{Gaudin, T.},
  \bibinfo{author}{Lanyi, D.}, \bibinfo{author}{Bekas, C.} \&
  \bibinfo{author}{Laino, T.}
\newblock \bibinfo{title}{“found in translation”: predicting outcomes of
  complex organic chemistry reactions using neural sequence-to-sequence
  models}.
\newblock \emph{\bibinfo{journal}{Chem. Sci.}}
  \textbf{\bibinfo{volume}{9}}~(28), \bibinfo{pages}{6091--6098}
  (\bibinfo{year}{2018}).

\bibitem{schwaller2019molecular}
\bibinfo{author}{Schwaller, P.} \emph{et~al.}
\newblock \bibinfo{title}{Molecular transformer: a model for
  uncertainty-calibrated chemical reaction prediction}.
\newblock \emph{\bibinfo{journal}{ACS Cent. Sci.}}
  \textbf{\bibinfo{volume}{5}}~(9), \bibinfo{pages}{1572--1583}
  (\bibinfo{year}{2019}).

\bibitem{o2012towards}
\bibinfo{author}{O’Boyle, N.~M.}
\newblock \bibinfo{title}{Towards a universal smiles representation-a standard
  method to generate canonical smiles based on the inchi}.
\newblock \emph{\bibinfo{journal}{J. Cheminf.}}
  \textbf{\bibinfo{volume}{4}}~(1), \bibinfo{pages}{1--14}
  (\bibinfo{year}{2012}).

\bibitem{schneider2015get}
\bibinfo{author}{Schneider, N.}, \bibinfo{author}{Sayle, R.~A.} \&
  \bibinfo{author}{Landrum, G.~A.}
\newblock \bibinfo{title}{Get your atoms in order — an open-source
  implementation of a novel and robust molecular canonicalization algorithm}.
\newblock \emph{\bibinfo{journal}{J. Chem. Inf. Model.}}
  \textbf{\bibinfo{volume}{55}}~(10), \bibinfo{pages}{2111--2120}
  (\bibinfo{year}{2015}).

\bibitem{pu2016variational}
\bibinfo{author}{Pu, Y.} \emph{et~al.}
\newblock \bibinfo{title}{Variational autoencoder for deep learning of images,
  labels and captions}.
\newblock In \emph{\bibinfo{booktitle}{Advances in Neural Information
  Processing Systems}}  (\bibinfo{publisher}{Curran Associates, Inc.},
  \bibinfo{year}{2016}).

\bibitem{he2021masked}
\bibinfo{author}{He, K.} \emph{et~al.}
\newblock \bibinfo{title}{Masked autoencoders are scalable vision learners}
  (\bibinfo{year}{2021}).

\bibitem{sun2021towards}
\bibinfo{author}{Sun, R.}, \bibinfo{author}{Dai, H.}, \bibinfo{author}{Li, L.},
  \bibinfo{author}{Kearnes, S.} \& \bibinfo{author}{Dai, B.}
\newblock \bibinfo{title}{Towards understanding retrosynthesis by energy-based
  models}.
\newblock In \emph{\bibinfo{booktitle}{Advances in Neural Information
  Processing Systems}} \bibinfo{pages}{10186--10194}
  (\bibinfo{publisher}{Curran Associates, Inc.}, \bibinfo{year}{2021}).

\bibitem{schneider2016s}
\bibinfo{author}{Schneider, N.}, \bibinfo{author}{Stiefl, N.} \&
  \bibinfo{author}{Landrum, G.~A.}
\newblock \bibinfo{title}{What’s what: The (nearly) definitive guide to
  reaction role assignment}.
\newblock \emph{\bibinfo{journal}{J. Chem. Inf. Model.}}
  \textbf{\bibinfo{volume}{56}}~(12), \bibinfo{pages}{2336--2346}
  (\bibinfo{year}{2016}).

\bibitem{jin2017predicting}
\bibinfo{author}{Jin, W.}, \bibinfo{author}{Coley, C.},
  \bibinfo{author}{Barzilay, R.} \& \bibinfo{author}{Jaakkola, T.}
\newblock \bibinfo{title}{Predicting organic reaction outcomes with
  weisfeiler-lehman network}.
\newblock In \emph{\bibinfo{booktitle}{Advances in Neural Information
  Processing Systems}}  (\bibinfo{publisher}{Curran Associates, Inc.},
  \bibinfo{year}{2017}).

\bibitem{englert2015efficient}
\bibinfo{author}{Englert, P.} \& \bibinfo{author}{Kov{\'a}cs, P.}
\newblock \bibinfo{title}{Efficient heuristics for maximum common substructure
  search}.
\newblock \emph{\bibinfo{journal}{J. Chem. Inf. Model.}}
  \textbf{\bibinfo{volume}{55}}~(5), \bibinfo{pages}{941--955}
  (\bibinfo{year}{2015}).

\bibitem{vaswani2017attention}
\bibinfo{author}{Vaswani, A.} \emph{et~al.}
\newblock \bibinfo{title}{Attention is all you need}.
\newblock In \emph{\bibinfo{booktitle}{Advances in Neural Information
  Processing Systems}}  (\bibinfo{publisher}{Curran Associates, Inc.},
  \bibinfo{year}{2017}).

\bibitem{irwin2022chemformer}
\bibinfo{author}{Irwin, R.}, \bibinfo{author}{Dimitriadis, S.},
  \bibinfo{author}{He, J.} \& \bibinfo{author}{Bjerrum, E.~J.}
\newblock \bibinfo{title}{Chemformer: a pre-trained transformer for
  computational chemistry}.
\newblock \emph{\bibinfo{journal}{Machine Learning: Science and Technology}}
  \textbf{\bibinfo{volume}{3}}~(1), \bibinfo{pages}{015022}
  (\bibinfo{year}{2022}).

\bibitem{sumner2020levenshtein}
\bibinfo{author}{Sumner, D.}, \bibinfo{author}{He, J.},
  \bibinfo{author}{Thakkar, A.}, \bibinfo{author}{Engkvist, O.} \&
  \bibinfo{author}{Bjerrum, E.~J.}
\newblock \bibinfo{title}{Levenshtein augmentation improves performance of
  smiles based deep-learning synthesis prediction} (\bibinfo{year}{2020}).

\bibitem{ucak2022retrosynthetic}
\bibinfo{author}{Ucak, U.~V.}, \bibinfo{author}{Ashyrmamatov, I.},
  \bibinfo{author}{Ko, J.} \& \bibinfo{author}{Lee, J.}
\newblock \bibinfo{title}{Retrosynthetic reaction pathway prediction through
  neural machine translation of atomic environments}.
\newblock \emph{\bibinfo{journal}{Nat. Commun.}}
  \textbf{\bibinfo{volume}{13}}~(1), \bibinfo{pages}{1--10}
  (\bibinfo{year}{2022}).

\bibitem{2016The}
\bibinfo{author}{Cao, Q.~M.}, \bibinfo{author}{Ma, X.~L.},
  \bibinfo{author}{Xiong, J.~M.}, \bibinfo{author}{Guo, P.} \&
  \bibinfo{author}{Chao, J.~P.}
\newblock \bibinfo{title}{The preparation of febuxostat by suzuki reaction}.
\newblock \emph{\bibinfo{journal}{Chin. J. New Drugs}}  (\bibinfo{year}{2016}).

\bibitem{guo2011enantioselective}
\bibinfo{author}{Guo, Z.-L.}, \bibinfo{author}{Deng, Y.-Q.},
  \bibinfo{author}{Zhong, S.} \& \bibinfo{author}{Lu, G.}
\newblock \bibinfo{title}{Enantioselective synthesis of (r)-salmeterol
  employing an asymmetric henry reaction as the key step}.
\newblock \emph{\bibinfo{journal}{Tetrahedron: Asymmetry}}
  \textbf{\bibinfo{volume}{22}}~(13), \bibinfo{pages}{1395--1399}
  (\bibinfo{year}{2011}).

\bibitem{nirogi2015design}
\bibinfo{author}{Nirogi, R.~V.}, \bibinfo{author}{Badange, R.},
  \bibinfo{author}{Reballi, V.} \& \bibinfo{author}{Khagga, M.}
\newblock \bibinfo{title}{Design, synthesis and biological evaluation of novel
  benzopyran sulfonamide derivatives as 5-ht6 receptor ligands}.
\newblock \emph{\bibinfo{journal}{Asian J. Chem.}}
  \textbf{\bibinfo{volume}{27}}~(6), \bibinfo{pages}{2117}
  (\bibinfo{year}{2015}).

\bibitem{ishiyama1995palladium}
\bibinfo{author}{Ishiyama, T.}, \bibinfo{author}{Murata, M.} \&
  \bibinfo{author}{Miyaura, N.}
\newblock \bibinfo{title}{Palladium (0)-catalyzed cross-coupling reaction of
  alkoxydiboron with haloarenes: a direct procedure for arylboronic esters}.
\newblock \emph{\bibinfo{journal}{J. Org. Chem.}}
  \textbf{\bibinfo{volume}{60}}~(23), \bibinfo{pages}{7508--7510}
  (\bibinfo{year}{1995}).

\bibitem{urawa2002investigations}
\bibinfo{author}{Urawa, Y.}, \bibinfo{author}{Naka, H.},
  \bibinfo{author}{Miyazawa, M.}, \bibinfo{author}{Souda, S.} \&
  \bibinfo{author}{Ogura, K.}
\newblock \bibinfo{title}{Investigations into the suzuki--miyaura coupling
  aiming at multikilogram synthesis of e2040 using (o-cyanophenyl) boronic
  esters}.
\newblock \emph{\bibinfo{journal}{J. Organomet. Chem.}}
  \textbf{\bibinfo{volume}{653}}~(1-2), \bibinfo{pages}{269--278}
  (\bibinfo{year}{2002}).

\bibitem{schwaller2021extraction}
\bibinfo{author}{Schwaller, P.}, \bibinfo{author}{Hoover, B.},
  \bibinfo{author}{Reymond, J.-L.}, \bibinfo{author}{Strobelt, H.} \&
  \bibinfo{author}{Laino, T.}
\newblock \bibinfo{title}{Extraction of organic chemistry grammar from
  unsupervised learning of chemical reactions}.
\newblock \emph{\bibinfo{journal}{Science Advances}}
  \textbf{\bibinfo{volume}{7}}~(15), \bibinfo{pages}{eabe4166}
  (\bibinfo{year}{2021}).

\end{thebibliography}

% \clearpage

% \input{supply}

\end{document}

\title[]{Supporting Information for Root-aligned SMILES: A Tight Representation for Chemical Reaction Prediction}

\author[1]{\fnm{Zipeng} \sur{Zhong}}\email{zipengzhong@zju.edu,cn}
\author[2]{\fnm{Jie} \sur{Song}}\email{sjie@zju.edu,cn}
\author[2]{\fnm{Zunlei} \sur{Feng}}\email{zunleifeng@zju.edu,cn}
\author[3]{\fnm{Tiantao} \sur{Liu}}\email{liutiant@zju.edu.cn}
\author[1]{\fnm{Lingxiang} \sur{Jia}}\email{lingxiangjia@zju.edu,cn}
\author[1]{\fnm{Shaolun} \sur{Yao}}\email{yaoshaolun@zju.edu,cn}
\author[4]{\fnm{Min} \sur{Wu}}\email{swgcwumin@hdnewdrug.com}
\author*[3]{\fnm{Tingjun} \sur{Hou}}\email{tingjunhou@zju.edu.cn}
\author*[1]{\fnm{Mingli} \sur{Song}}\email{brooksong@zju.edu.cn}

\affil[1]{\orgdiv{College of Computer Science and Technology}, \orgname{Zhejiang University}, \orgaddress{\postcode{310027}, \state{Zhejiang}, \country{P.R. China}}}

\affil[2]{\orgdiv{School of Software Technology}, \orgname{Zhejiang University}, \orgaddress{\postcode{315048}, \state{Zhejiang}, \country{P.R. China}}}

\affil[3]{\orgdiv{Innovation Institute for Artificial Intelligence in Medicine of Zhejiang University, College of Pharmaceutical Sciences}, \orgname{Zhejiang University}, \orgaddress{\postcode{310058}, \state{Zhejiang}, \country{P.R. China}}}

\affil[4]{ \orgname{Hangzhou Huadong Medicine Group Pharmaceutical Research Institute}, \orgaddress{\postcode{310011}, \state{Zhejiang}, \country{P.R. China}}}

\maketitle

\section{Additional Information for Methods}
\textbf{Vanilla transformer} We take the vanilla transformer as the backbone of our autoencoder. Vanilla transformer~\cite{vaswani2017attention} is an end-to-end model following a stepwise and autoregressive encoder-decoder fashion. Taking the product SMILES and partially decoded reactant SMILES as the input, it is trained to predict the next token of reactant SMILES.The key idea of the vanilla transformer is the attention mechanism, which allows each token to capture the global information and is quite suitable for SMILES representation. The encoder and decoder are both composed of multiple stacked multihead attention layers consisting of a multihead attention module and a position-wise feed forward module.

Before passing into the encoder, SMILES tokens are embedded to continuous vector representations. The multihead attention module consists of multiple scaled-dot product layers that run in parallel. A single scaled-dot product calculation works as follows:

\begin{equation}
    \begin{gathered}
    Attention(Q,K,V) = softmax(\frac{QK^T}{\sqrt{d_k}})V  \\
    Q = W_QX' \\
    K = W_KX' \\
    V = W_VX'
    \end{gathered}
\end{equation}
	 	
where $Q$, $K$, $V$ represent query, key, value matrix, respectively; $W_Q$, $W_K$, $W_V$ are all trainable parameters; $d_k$ means the dimension of $K$. Depending on where $Q$, $K$, $V$ come from, multihead attention can be the self-attention mechanism or cross-attention mechanism. After the attention calculation of each head, they can be concatenated as follows:

\begin{equation}
    \begin{gathered}
    Z = Concat(h_0,h_1,...)W^0 \\
    h_i = Attention(Q_i,K_i,V_i)
    \end{gathered}
\end{equation}
	 	
The position-wise feed forward module is a simple fully connected layer that utilizes the concept of residual block and works as follows:

\begin{equation}
    FFN(Z)=max(0,W_1z+b_1)W_2+b_2
\end{equation}
	 	
After the calculation of feed forward module, updated token vectors can be passed to another multihead attention layer. We use the vanilla transformer architecture composed of 6 layers for both encoder and decoder with 8 attention heads for all experiments.
During the inference stage, the transformer takes the product SMILES and decoded reactant as the input to predict the probability of the next reactant SMILES token, which can be represented by a conditional probability distribution:

\begin{equation}
    p(y\vert X) = \prod^{m}_{i=1}p(y_i\vert y_{<i},X)
\end{equation}
	 	
where $m$ is the maximum number of reactant tokens. The ``bos'' (begin of sentence) token is the beginning of reactant tokens. When the last predicted token is ``eos'' (end of sentence), the decoding process completes.

\begin{figure*}
    \centering
    \includegraphics[width=\textwidth]{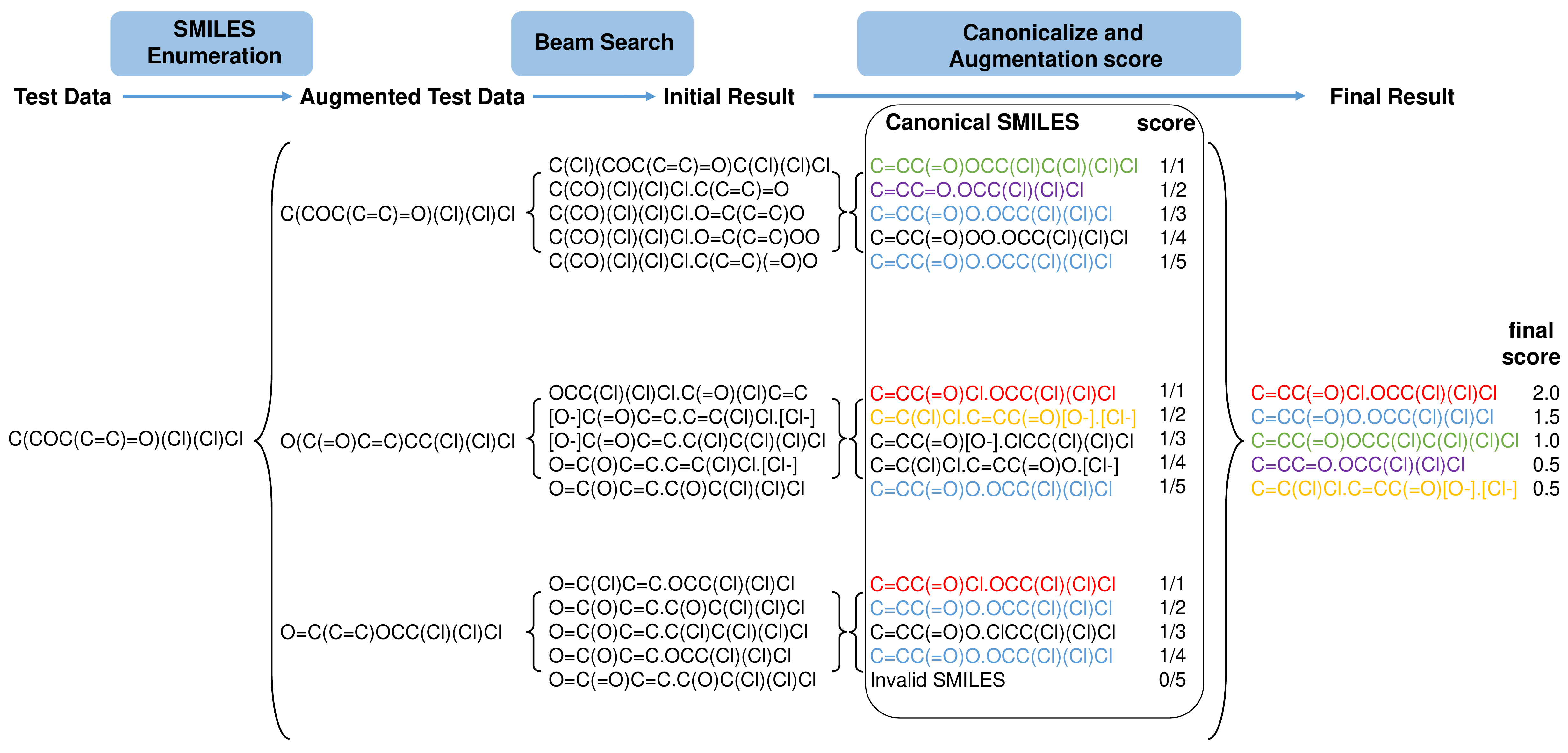}
    \caption{The workflow of model prediction with the beam search and data augmentation. The example applies $3 \times$ augmentation, and sets $beam\ size=5$, $topk=5$.}
    \label{fig:test_set_aug}
\end{figure*}

\noindent\textbf{Data augmentation with R-SMILES}
We successively perform data augmentation and root alignment on the training data, and only perform data augmentation on the test data. When inferring on the validation and test data, we input multiple SMILES of a molecule respectively and get multiple sets of outputs correspondingly.
After removing invalid SMILES that cannot be recognized by Rdkit~\cite{landrum2013rdkit} and converting all outputs to canonical SMILES, we refer to Tekto \textit{et al.}'s approach~\cite{tetko2020state} that scores these outputs uniformly as follows:
\begin{equation}
    \centering
    score(output)=\sum_{n=1}^{augmentation}\sum_{i=k}^{topk}\frac{1}{k}
    \label{rank_eq}
\end{equation}
	 	
\noindent where  $augmentation$  represents the augmentation times of the test set, and  $beam$ represents the beam size. After scoring uniformly, we can select outputs with top-K scores as the final result.

Here we show an example of performing data augmentation for the test data in Figure \ref{fig:test_set_aug}.
When the SMILES ``C(COC(C=C)=O)(Cl)(Cl)Cl'' was input, first we performed SMILES enumeration to get three different SMILES representing the same molecule. Then we started model prediction using the beam search strategy with the beam size of 5, and got the top-5 prediction for each SMILES. To score uniformly, we converted them to the canonical SMILES and removed those invalid SMILES. According to the ranking of the initial results, we can give each output an initial score $\frac{1.0}{k}$. For example, the rank-1 prediction ``C=CC(=O)OCC(Cl)C(Cl)(Cl)Cl'' of ``C(COC(C=C)=O)(Cl)(Cl)Cl'' was scored one. If the prediction was an invalid SMILES, we would score it zero. After getting the scores for each output, we can score them uniformly by adding the scores of the same output. For example, since ``C=CC(=O)Cl.OCC(Cl)(Cl)Cl'' was the rank-1 prediction for two inputs  ``O(C(=O)C=C)CC(Cl)(Cl)Cl'' and ``O=C(C=C)OCC(Cl)(Cl)Cl'', its final score was two. Therefore, we can acquire the scores for all the output and got a uniform ranking, i.e, the final result.

\begin{table*}[htbp]
  \centering
  \caption{The effect of choosing different values of $\alpha$ on the top-K single-step retrosynthesis.}
  \resizebox{\textwidth}{!}{
\begin{tabular}{ccccccccccccc}
\hline
      & \multicolumn{4}{c}{\textbf{USPTO-50K top-K Accuracy (\%)}} & \multicolumn{4}{c}{\textbf{USPTO-MIT top-K Accuracy (\%)}} & \multicolumn{4}{c}{\textbf{USPTO-FULL top-K Accuracy (\%)}} \\
\hline
$\alpha$ & K = 1 & 3     & 5     & 10    & K = 1 & 3     & 5     & 10    & K = 1 & 3     & 5     & 10 \\
\hline
0.001 & 54.2  & 75.7  & 85.0  & 89.5  & 59.4  & 76.0  & 81.3  & 86.0  & 47.4  & 63.5  & 69.5  & 74.8  \\
0.01  & 54.1  & 75.7  & 85.1  & 89.5  & 58.8  & 74.7  & 79.2  & 84.3  & 47.8  & 63.7  & 68.3  & 73.4  \\
0.1   & 55.1  & 77.6  & 84.0  & 90.0  & 59.3  & 76.6  & 81.9  & 86.4  & 48.8  & 64.9  & 70.1  & 76.0  \\
1     & \textbf{56.0 } & \textbf{79.4 } & \textbf{86.2 } & \textbf{91.2 } & \textbf{60.4 } & \textbf{78.4 } & 83.4  & \textbf{87.6 } & \textbf{49.0 } & 66.4  & 71.8  & \textbf{76.4 } \\
10    & 55.9  & 79.2  & 86.1  & 91.1  & 60.2  & \textbf{78.4 } & \textbf{83.5 } & 87.5  & 48.9  & \textbf{66.5 } & 71.6  & \textbf{76.4 } \\
100   & 55.9  & 79.1  & \textbf{86.2 } & \textbf{91.2 } & 60.3  & 78.2  & 83.4  & 87.5  & 48.9  & \textbf{66.5 } & \textbf{71.9 } & \textbf{76.4 } \\
\hline
\end{tabular}%

    }
  \label{table:alpha_effect}%
\end{table*}%

There is a scoring trade-off in the test set data augmentation, that is, how to weigh the number of prediction occurrences against the ranking of predictions. 
Suppose there are two predictions, one that is predicted by one input and ranks first, and the other that is predicted by two different inputs and ranks second and third respectively, which of them should get a higher final score? 
This problem can be expressed by the following equation:
\begin{equation}
    \centering
    score(output)=\sum_{n=1}^{augmentation}\sum_{i=k}^{topk}\frac{1}{1 + \alpha * (k - 1)}
    \label{original_rank_eq}
\end{equation}

\noindent where $\alpha$ is the weighing parameter. The higher $\alpha$, the more important the ranking and the less important the number of appearances, and vice versa. We tested different values of $\alpha$ and show the accuracy of the validation set in the Table~\ref{table:alpha_effect}. 
It can be seen that for all datasets, the best results are obtained when the $\alpha$ value is equal to or greater than one, which demonstrates the ranking is more important than the number of occurrences. Moreover, when the $\alpha$ value is one, most of the top-K accuracies are the highest.
In fact, Eq.~\ref{rank_eq} is the case where $\alpha$ takes the value of one in Eq.~\ref{original_rank_eq}.

\begin{table*}[htbp]
  \centering
  \caption{Training time and steps for different tasks. ``R2P'' denotes the forward reaction prediction, and  ``P2R'' denotes the retrosynthesis prediction. ``From Scratch'' denotes the model is trained without pretraining.}
  \resizebox{1.0\textwidth}{!}{
% Table generated by Excel2LaTeX from sheet 'Sheet6'
\begin{tabular}{cccc}
\hline
\textbf{Dataset} & \textbf{USPTO-50K} & \textbf{USPTO-MIT} & \textbf{USPTO-Full} \\
\hline
Pretrain & - & - & 130 hours / 1,000,000 steps \\
Finetune - R2P & - & 30 hours / 500, 000 steps & - \\
Finetune - P2R & 20 hours / 300,000 steps & 50 hours / 500,000 steps & 50 hours / 500,000 steps \\
From Scratch - R2P & - & 60 hours / 100,000 steps & - \\
From Scratch - P2R & 30 hours / 600,000 steps & 110 hours / 2,000,000 steps & 120 hours / 1,500,000 steps \\
\hline
\end{tabular}%
}
\label{table:timing}%
\end{table*}%

\noindent\textbf{Training settings} 
We use the same data split as previous researchers~\cite{coley2017computer, jin2017predicting,dai2019retrosynthesis} for all the datasets.
During the pretraining stage. Depending on whether it is a forward or retrosynthesis prediction, products or reacatants in the training set of USPTO-FULL are used for self-supervised training, where molecules in the test set of USPTO-50K and USPTO-MIT are removed. 
We apply $20\times$ augmentation at training and test sets of USPT0-50K, and $5\times$ augmentation at training and test sets of USPTO-MIT and USPTO-FULL. We set the embedding and hidden size as 512 except that the dimension of  $Q$ , $K$ , $V$  is 64. We also use the Adam optimizer and a varied learning rate with 8,000 warm up steps. The input and output share the same vocabulary, but their embedding layers are independent. 
We conducted all experiments with one NVIDIA GeForce RTX 3090 GPU, and the approximate training time and steps for different tasks are displayed in Fig.~\ref{table:timing}.
The more detailed settings can be found at https://github.com/otori-bird/retrosynthesis.

\begin{figure}
    \centering
    \includegraphics[width=0.45\textwidth]{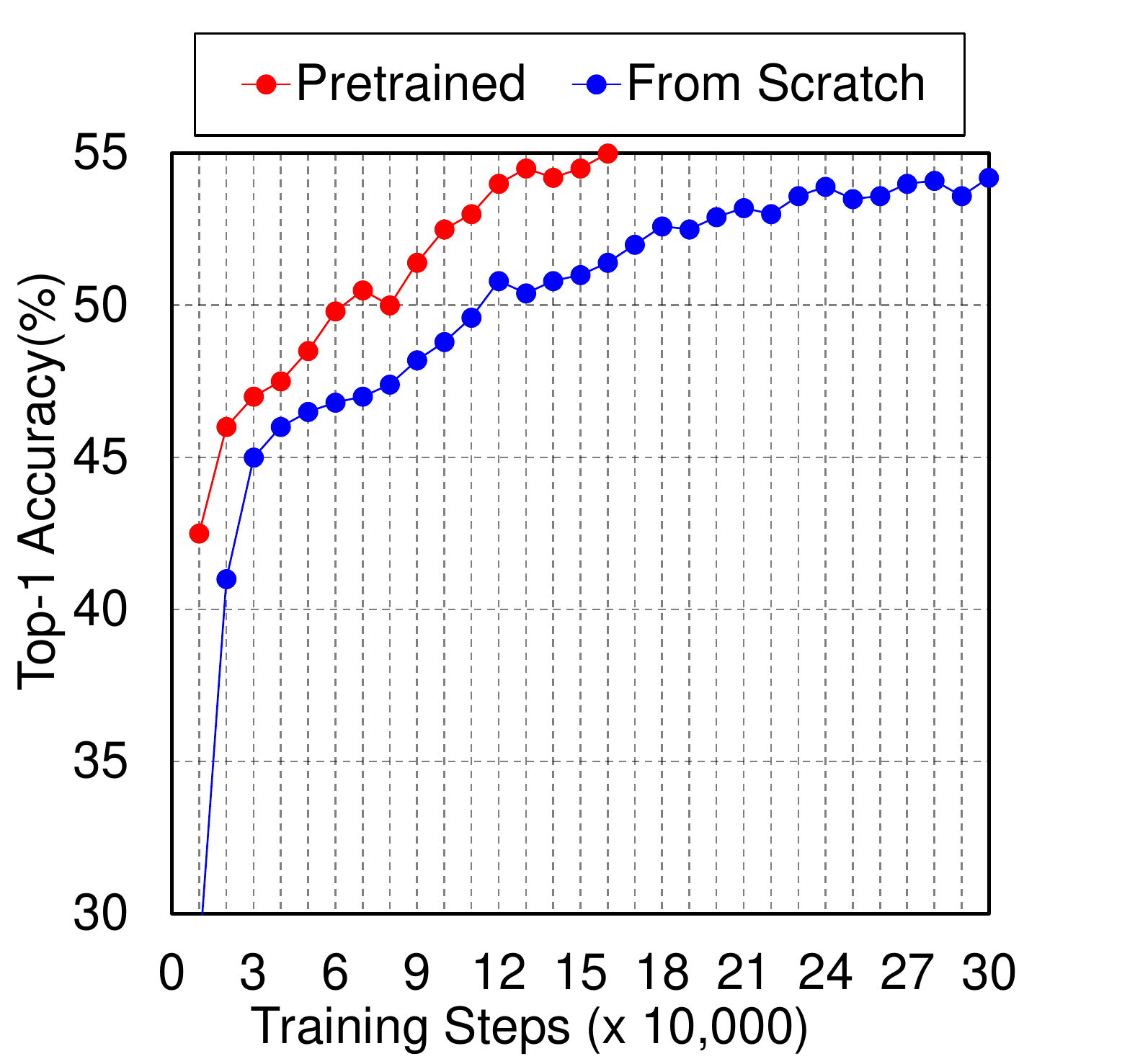}
    \caption{Training steps with/without the pretrained model on USPTO-50K for the P2R stage. The training set of USPTO-50K is applied 20$\times$ augmentation. }
    \label{fig:pretrain_time}
\end{figure}

\section{Analysis of Pretrained Transformer}
We used a similar approach to the masked language model of BERT~\cite{devlin2018bert} for pretraining. Specifically, 15\% of tokens in SMILES are masked. Every masked token has an 80\% probability of being replaced with the ``unknown'' token, a 10\% probability of being replaced with any token in the vocabulary, and keeps unchanged for the rest of the cases. After pretraining, we can see in Figure~\ref{fig:pretrain_time} and Table~\ref{table:timing} that the training time has been dramatically reduced, which helps a lot in the case of very limited computational resources.

\begin{table*}[htbp]
  \centering
  \caption{The edit distance and top-K accuracy of single-step retrosynthesis for ring and non-ring reactions on the USPTO-MIT and USPTO-FULL datasets.}
  \resizebox{1.0\textwidth}{!}{
    \begin{tabular}{ccccccccccc}
    \hline
          & \multicolumn{5}{c}{\textbf{USPTO-MIT}} & \multicolumn{5}{c}{\textbf{USPTO-FULL}} \\
    \hline
    \textbf{Reaction Type} & \textbf{edit distance} & \textbf{K=1} & \textbf{3} & \textbf{5} & \textbf{10} & \textbf{edit distance} & \textbf{K=1} & \textbf{3} & \textbf{5} & \textbf{10} \\
    \hline
    Overall\textsuperscript{\emph{a}} & 26.7  & 53.8  & 70.4  & 75.1  & 77.7  & 29.2  & 44.7  & 61.0  & 65.8  & 68.7  \\
    Non-ring reaction\textsuperscript{\emph{a}} & 25.9  & 55.9  & 73.6  & 78.5  & 81.3  & 27.3  & 49.6  & 67.7  & 72.9  & 76.0  \\
    Ring-opening reaction\textsuperscript{\emph{a}} & 33.5  & 39.7  & 48.0  & 50.4  & 52.0  & 39.1  & 25.3  & 33.3  & 35.8  & 37.8  \\
    Ring-forming reaction\textsuperscript{\emph{a}} & 23.7  & 37.9  & 51.6  & 56.1  & 58.7  & 28.4  & 34.1  & 49.2  & 54.3  & 57.5  \\
    \hline
    Overall\textsuperscript{\emph{b}} & 13.5 (-49\%) & 60.4  & 78.0  & 83.0  & 86.9  & 16.6 (-43\%) & 48.9  & 66.5  & 71.8  & 76.7  \\
    Non-ring reaction\textsuperscript{\emph{b}} & 12.5 (-52\%) & 62.8  & 80.5  & 85.3  & 89.4  & 13.9 (-49\%) & 54.1  & 72.6  & 77.9  & 83.0  \\
    Ring-opening reaction\textsuperscript{\emph{b}} & 21.8 (-35\%) & 43.3  & 57.2  & 62.9  & 67.2  & 28.4 (-28\%) & 27.3  & 39.1  & 43.9  & 48.8  \\
    Ring-forming reaction\textsuperscript{\emph{b}} & 15.2 (-36\%) & 46.8  & 63.8  & 70.4  & 76.8  & 20.8 (-27\%) & 39.3  & 56.7  & 63.2  & 69.6  \\
    \hline
    \end{tabular}%
    }
    
\textsuperscript{\emph{a}} Without root alignment;
\textsuperscript{\emph{b}} With root alignment.
  \label{tab:ring_acc}%
\end{table*}%

\section{Limitations for ring-opening and ring-forming reactions}
The root alignment strategy does not always work well. Taking the ring-opening reaction 
``[CH2:1]1[CH:2]([NH2:3])[CH:4]([OH:5])[O:6][CH2:7]1
\noindent$>>$[CH2:1]([CH:2]([NH2:3])[CH:4]=[O:5])[CH2:7]'' as an example, where the bond between ``[O:6]'' and ``[CH:4]'' is split, we can get both the reasonable aligned result and largely unaligned one. Aligning at the root atom ``[O:6]'' yields reasonable results ``O1CCC(N)C1O'' and ``OCCC(N)C=O'', whereas aligning at root atom ``[O:5]'' yields largely unaligned strings ``OC1OCCC1N'' and ``O=CC(N)CCO''. This obviously increases the edit distance between inputs and outputs, which leads to a decrease in prediction performance for this type of data.We calculated the accuracy of retrosynthesis for ring-opening and forming reactions in different datasets. Results are shown in Table~\ref{tab:ring_acc}. It can be seen that the accuracy of R-SMILES, as pointed out by the reviewer, is not so high as that of other reactions. To make it clearer, we also calculated the edit distance between the input and the output SMILES for these reactions. Compared with that of non-ring reaction R-SMILES, the edit distance of ring reactions is significantly larger. These results again verify our main motivation in this work that large distance between input and output strings will degrade the reaction prediction performance. We will be devoted to dealing with this problem in our future work by trying to align more than one atom.

\begin{figure}
    \centering
    \includegraphics[width=.5\textwidth]{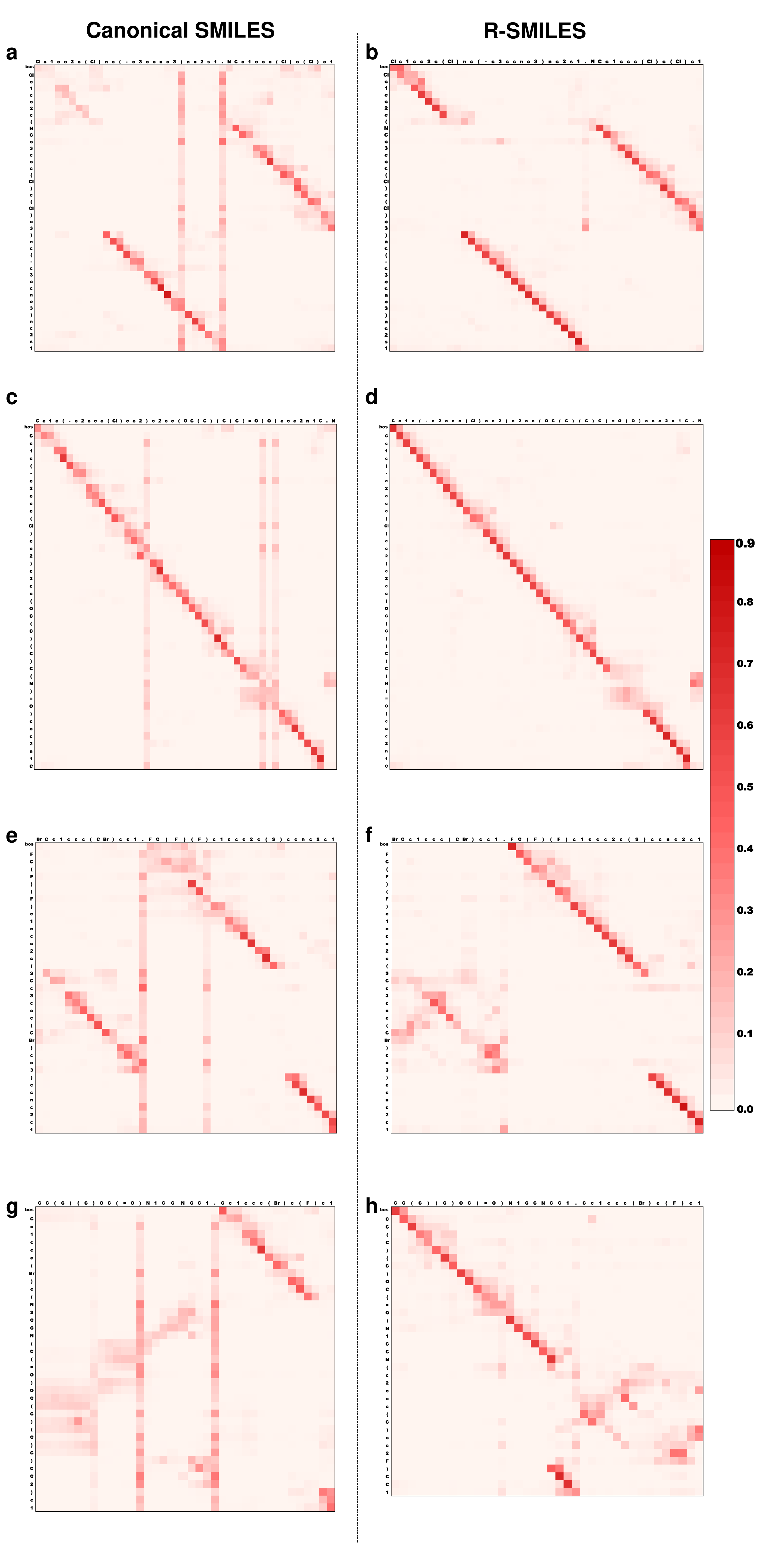}
    \caption{Visualization of the cross-attention obtained by the canonical SMILES~(Left) and the proposed R-SMILES~(Right) in the forward reaction prediction.
    (a, c, e, g) The attention maps obtained by the model trained with canonical SMILES.
    (b, d, f, h) The attention maps obtained by the model trained with R-SMILES.
     The input tokens are along the x axis, and the output tokens are along the y axis.
     Each row in the attention map represents the attention over the input tokens for predicting the next output token.
     Each column represents the attention between an input token with each output token.
     The ``bos'' token is the beginning of output tokens and will be removed after the decoding process completes.
     }
    \label{fig:attention_1}
\end{figure}

\begin{figure*}
    \centering
    \includegraphics[width=\textwidth]{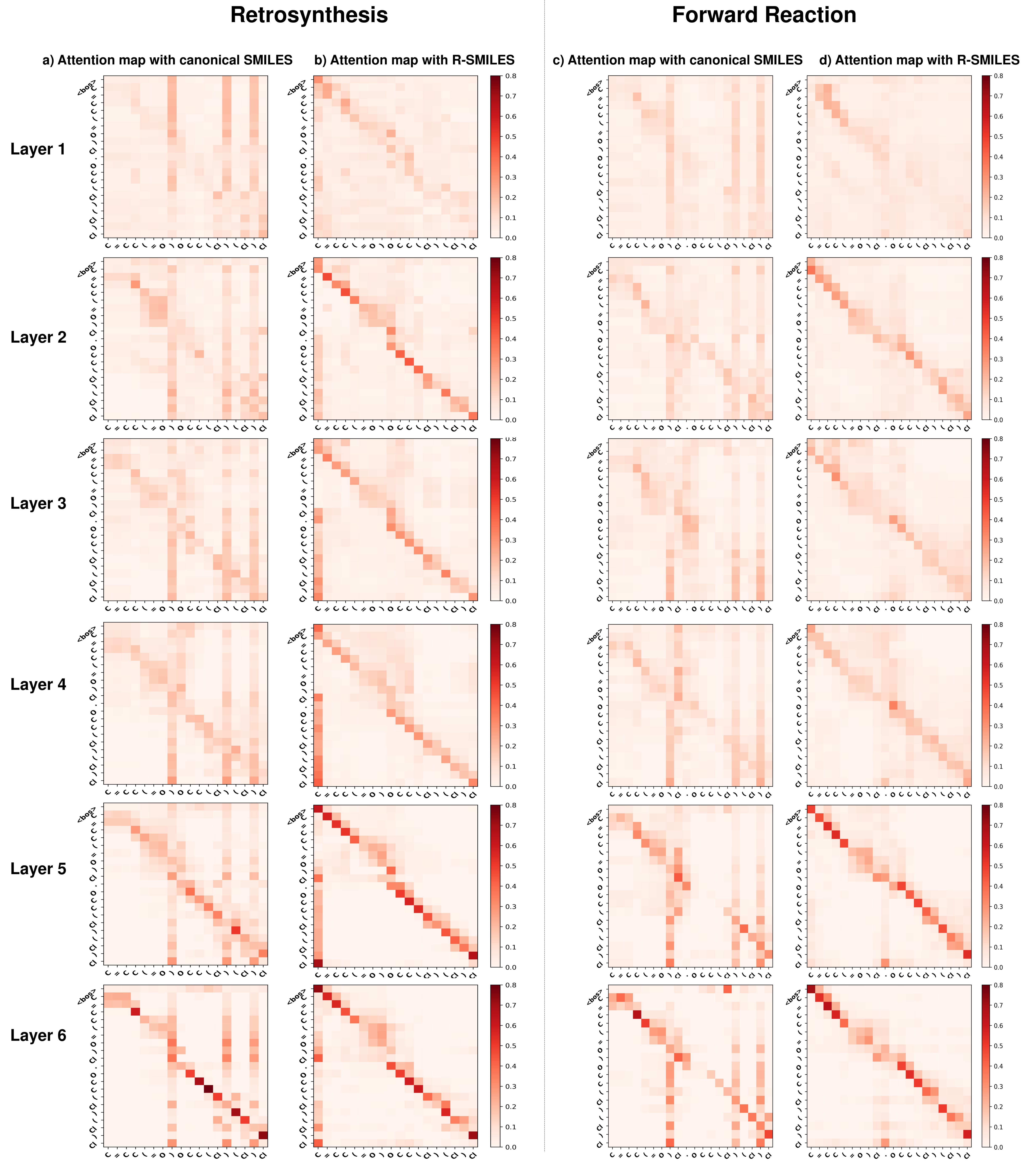}
    \caption{Comparison of the attention of the all the layers.
    (a, c) are the attention maps obtained by the model trained with canonical SMILES.
    (b, d) are the attention maps obtained by the model trained with R-SMILES.}
    \label{fig:attention_2}
\end{figure*}

\section{Comparison of the attention of all layers}
To give a more detailed discussion about attention mechanism, we fed the reactant or the product of the reaction ``C=CC(=O)Cl.OCC(Cl)(Cl)Cl$>>$C=CC(=O)OC
\noindent C(Cl)(Cl)Cl''  to four different models and visualized attention maps of all layers in Fig.~\ref{fig:attention_2}. For these four models, one is trained with R-SMILES for forward reaction, one trained with R-SMILES for retrosynthesis, one trained with canonical SMILES for forward reaction, and one trained with SMILES for retrosynthesis.
With SMILES~(Fig.~~\ref{fig:attention_2}a, c), much attention is paid to syntactic tokens at the shallow layers.
However, for R-SMILES, the attention maps are always cleaner and nearly diagonal at different layers. Attention maps of the 200 examples can be found at https://github.com/otori-bird/retrosynthesis.

\begin{figure*}
    \centering
    \includegraphics[width=\textwidth]{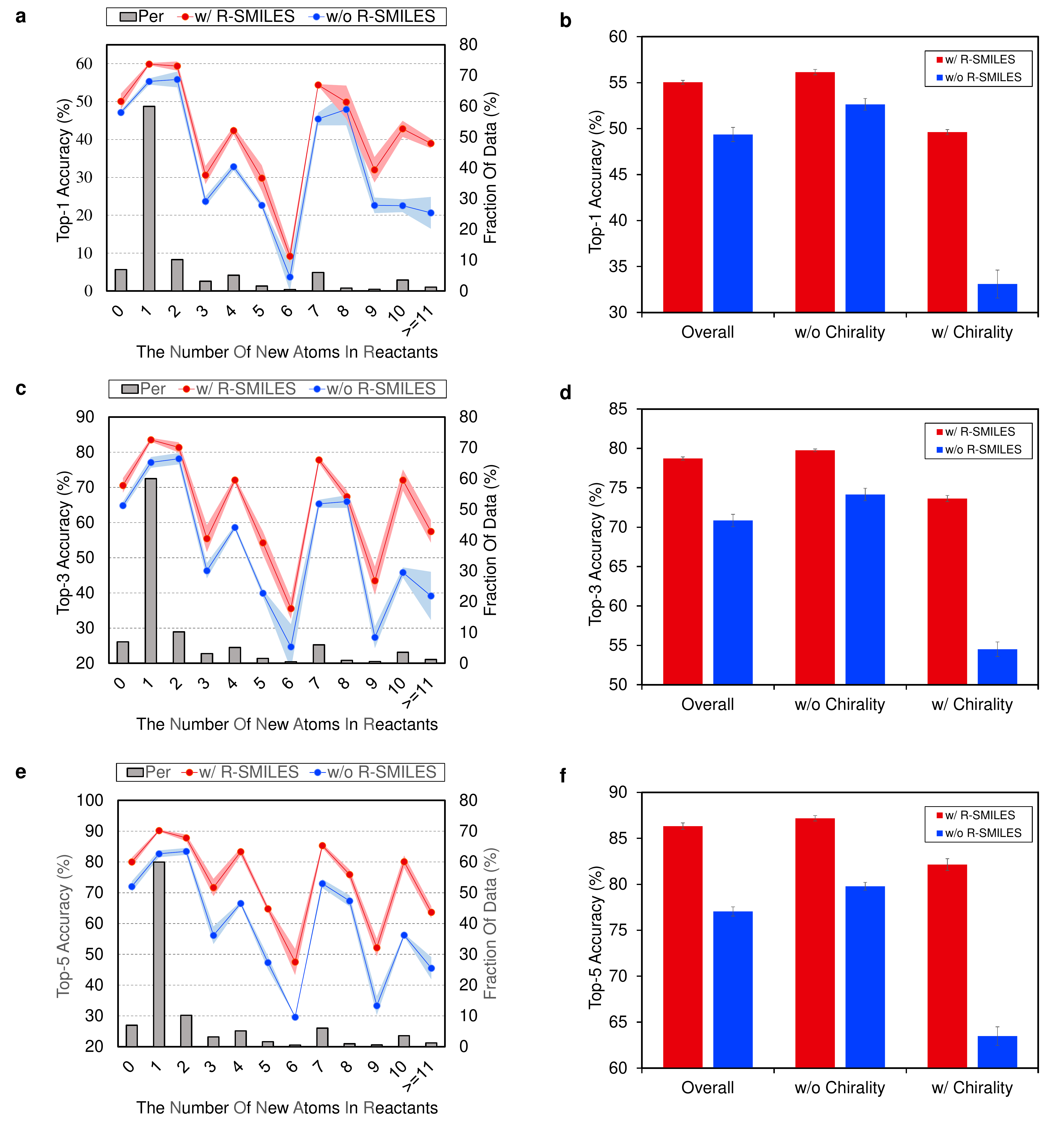}
    \caption{Extra top-K accuracy (\%) for complex reactions.  \textbf{a, c, e,} Top-1, top-3, and top-5 accuracies according to the number of new atoms in reactants. The red and blue lines represent the performance with/without R-SMILES. The gray bar means the percentage of this kind of reaction in the test set. \textbf{b, d, f,}  Top-1, top-3, and top-5 accuracies for reactions involving with or without chirality. The red and blue bars represent the performance with or without R-SMILES.}
    \label{fig:attention_3}
\end{figure*}
\clearpage

% \bibliography{sn-article}% common bib file